\documentclass{article}

\usepackage{arxiv}

\usepackage[utf8]{inputenc} 
\usepackage[T1]{fontenc}    
\usepackage{hyperref}       
\usepackage{url}            
\usepackage{booktabs}       
\usepackage{amsfonts}       
\usepackage{nicefrac}       
\usepackage{microtype}      
\usepackage{graphicx}
\usepackage{doi}

\usepackage{tabu}                      
\usepackage{amsmath}
\usepackage{mathtools}
\usepackage{multirow}
\usepackage{enumitem}

\ifpdf
  \pdfoutput=1\relax                   
  \usepackage{graphicx}                
  \DeclareGraphicsExtensions{.pdf,.png,.jpg,.jpeg} 
\else
  \usepackage{graphicx}                
  \DeclareGraphicsExtensions{.eps}     
\fi%

\usepackage{xcolor}
\definecolor{cb_orange}{rgb}{1.0,0.51,0.0}
\definecolor{cb_blue}{rgb}{0.22,0.49,0.72}
\definecolor{cb_green}{rgb}{0.3,0.67,0.29}
\definecolor{cb_red}{rgb}{0.89,0.1,0.11}
\definecolor{cb_purple}{rgb}{0.6, 0.31, 0.64}
\definecolor{cadetgrey}{rgb}{0.57, 0.64, 0.69}

\title{MitoVis: A Visually-guided Interactive Intelligent System for Neuronal Mitochondria Analysis}

\author{ JunYoung Choi, Hakjun Lee, Suyeon Kim, Seok-Kyu Kwon, and Won-Ki Jeong}

\renewcommand{\headeright}{}
\renewcommand{\undertitle}{}
\renewcommand{\shorttitle}{Choi et al.: MitoVis}

\hypersetup{
pdftitle={Choi et al. MitoVis},
pdfsubject={.},
pdfauthor={JunYoung Choi},
pdfkeywords={Biomedical and Medical Visualization, Machine Learning, Task and Requirements Analysis, User Interfaces, Intelligence Analysis},
}

\begin{document}
\maketitle

\begin{abstract}

Neurons have a polarized structure, including dendrites and axons, and compartment-specific functions can be affected by dwelling mitochondria. 
It is known that the morphology of mitochondria is closely related to the functions of neurons and neurodegenerative diseases. 
Even though several deep learning methods have been developed to automatically analyze the morphology of mitochondria, the application of existing methods to actual analysis still encounters several difficulties. 
Since the performance of pre-trained deep learning model may vary depending on the target data, re-training of the model is often required. 
Besides, even though deep learning has shown superior performance under a constrained setup, there are always errors that need to be corrected by humans in real analysis. 
To address these issues, we introduce MitoVis, a novel visualization system for end-to-end data processing and interactive analysis of the morphology of neuronal mitochondria. 
MitoVis enables interactive fine-tuning of a pre-trained neural network model without the domain knowledge of machine learning, which allows neuroscientists to easily leverage deep learning in their research.
MitoVis also provides novel visual guides and interactive proofreading functions so that the users can quickly identify and correct errors in the result with minimal effort.
We demonstrate the usefulness and efficacy of the system via a case study conducted by a neuroscientist on a real analysis scenario. 
The result shows that MitoVis allows up to 15$\times$ faster analysis with similar accuracy compared to the fully manual analysis method.


\end{abstract}

\keywords{Biomedical and Medical Visualization \and Machine Learning \and Task and Requirements Analysis \and User Interfaces \and Intelligence Analysis}

\begin{figure}[b]
  \centering
  \includegraphics[width=\linewidth]{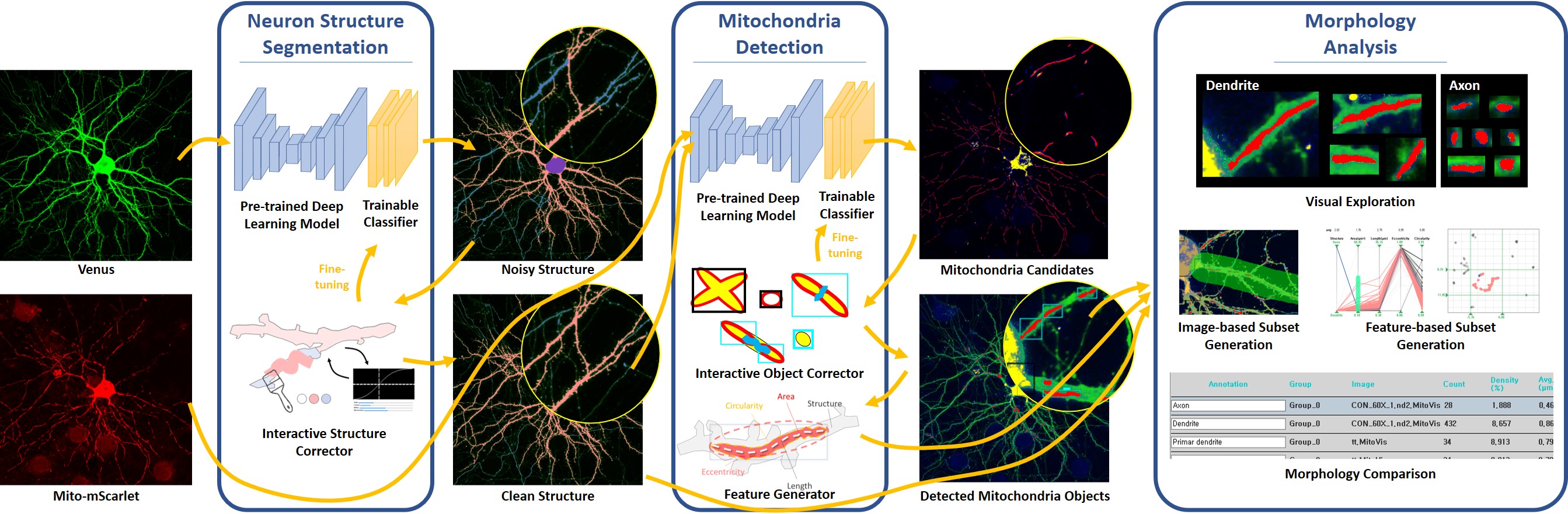}
  \caption{Overview of MitoVis. MitoVis is a visual analytics system for neuronal mitochondria analysis with effective visualizations and interactive deep learning. The yellow lines represent data flow.}
  \label{fig:teaser}
\end{figure}

\section{Introduction}

Neurons are unique cells that have highly polarized structures, including dendrites and axons, and perform specialized functions for synaptic transmission. More interestingly, one of the intracellular organelles, mitochondria, also displays compartmentalized features in dendrites and axons. Several studies have reported that dendritic mitochondria have long and tubular shapes, whereas axonal mitochondria exhibit short and punctate shapes~\cite{chicurel1992three,dickey2011pka,kasthuri2015saturated,li2004importance,popov2005mitochondria,lewis2018mff}. The difference in mitochondrial morphology is also associated with the function of each compartment such as neurotransmitter release and dendritic protein synthesis via calcium clearance and ATP generation~\cite{lewis2018mff,rangaraju2019spatially,rangaraju2019pleiotropic}. Additionally, in neurodegenerative disease models, including Alzheimer’s disease and Parkinson’s disease, it have been widely studied that dendritic mitochondria are fragmented~\cite{schon2011mitochondria,wang2009impaired,lee2018emerging,baek2017inhibition,zahedi2018deep}. Therefore, analyzing the morphology of mitochondria in each compartment is critical for both normal neuronal function and disease models.

To analyze the morphology of dendritic and axonal mitochondria, several pre-processing steps, such as segmenting dendrites and axons from neuron images, detecting mitochondria, and extracting morphological features, are required. 
These tasks have been mainly performed by manually annotating dendritic mitochondria and axonal mitochondria, which is time-consuming and labor-intensive for large datasets. 
%
%
Some recent studies introduced deep-learning-based algorithms for automated neuronal structure segmentation and mitochondria detection. 
For example, Fischer et al.~\cite{fischer2020mitosegnet} showed that a deep learning model can be used to achieve a higher segmentation accuracy of mitochondria than that achieved by conventional feature-based segmentation methods. 
Park et al.~\cite{chan2021noiseloss} proposed novel loss functions to train a deep neural network for the segmentation of cell bodies, dendrites, and axons from neuron images using noisy and incomplete training labels. 
%


Although these methods have paved the way for new research directions toward automated mitochondria analysis, the application of existing methods for actual analysis still encounters several difficulties. 
First, even though deep learning has delivered superior performance in previous studies, its actual performance may vary depending on the target data presented at its deployment; therefore, re-training of the model for the actual target data is often required. 
This is typically a time-consuming and difficult task for biologists, which hinders the adaptation of such deep-learning methods for their day-to-day analysis workflow.
Second, even though the deep learning method yields good results overall, there are always errors that need to be corrected by humans. 
However, there is no existing system that can effectively proofread segmented neuron structures and detected mitochondria. 
%
%
Third, the existing mitochondria analysis workflow relies on a loose combination of several stand-alone programs, and there is no unified system for realizing end-to-end data processing and analysis specifically designed for neuronal mitochondria analysis tasks. 

To address these issues, we propose a visual analysis system, \emph{MitoVis}, for realizing neuronal mitochondrial morphology analysis (such as neuronal structure segmentation, mitochondria detection, morphological feature extraction, and comparative analysis). 
MitoVis allows fast and accurate neuronal structure segmentation and mitochondria detection via interactive fine-tuning of the deep learning model through novel visual guides and intuitive user interactions. 
With this approach, users can easily leverage recent deep learning methods in their workflow using only simple user interactions at the visual interface.
%
%
Subset generation functions enable the comparative analysis of morphologies by creating various mitochondria subsets.
Additionally, MitoVis provides a unified system for end-to-end data processing and analysis workflow specifically designed to address the needs of neuroscience researchers, which allows significantly faster analysis with fewer user interactions for large-scale datasets (up to 15$\times$ faster than the conventional workflow).
We demonstrate the performance and usability of this system using real-world analysis.
%
%
%
%
Our main contributions are summarized as follows:
\begin{itemize}
    \item We propose a unified visual analysis system that can perform all the necessary data processing and analysis tasks for neuronal mitochondria analysis through effective visualization and deep learning.
    \item We propose an interactive deep learning approach that allows non-experts on deep learning to perform neuronal structure segmentation and mitochondria detection precisely through deep learning with visual guides and intuitive interactions.
    \item We make the proposed software system (and the related source code) open to the public to enable the practical use of our research and the development of the research field.
\end{itemize}

\section{Background and Related Work}

\subsection{Mitochondria detection and morphology analysis}

The morphology of mitochondria using fluorescence images has been analyzed in several ways. The basic method is manual measurement using an image processing program such as commercially available software and an open-source program such as ImageJ~\cite{merrill2017measuring}. However, it is time-consuming and has limitations in handling a large dataset. Therefore, attempts have been made to develop automated analysis programs for mitochondrial morphology~\cite{zahedi2018deep,lihavainen2012mytoe,fischer2020mitosegnet}. These programs were applied for conducting mitochondrial analysis in non-neuronal cells; however, they are not suitable for neuronal mitochondria, which have a distinct morphology during the processes. To overcome these challenges, several groups have developed their own pipelines by integrating existing software, although these require multiple manual pre-processing steps and classify dendritic and axonal mitochondria depending on their length rather than the structure of neurites~\cite{varkuti2020neuron,lewis2018mff}. A recent machine learning-based mitochondrial classification program for neurons mainly focuses on the structure of mitochondria around the cell body and cannot distinguish between dendrites and axons~\cite{fogo2021machine}.

\subsection{Neuronal structure segmentation}

In general, neuronal dendrites display a thicker tubular shape than axons in the mammalian brain. To specify the neuronal compartments, a manual or semi-automatic approach has been applied using open-source or commercially available software~\cite{zhou2020gtree,feng2015neutube,peng2014extensible,magliaro2017manual}. However, these approaches require significant user effort and time.


Recently, one study~\cite{chan2021noiseloss} was conducted to segment dendrites, axons, and cell bodies from florescence images through deep learning. In this study, a new training loss is defined to tackle the difficult encountered in creating a precise training label from a neuronal image with a complex structure; precise segmentation can be performed with only partially labeled training data.


\subsection{Interactive deep learning}

Deep learning has been verified to deliver high performance and has been widely used in the image segmentation field. 
However, because the performance of supervised learning largely depends on the training data, 
it is often not possible to obtain accurate results for unseen data that are not included in the training data. 
Recently, an interactive learning approach that improves the performance by modifying the model through user interaction has been studied to overcome this limitation. 
For example, Wang et al.~\cite{wang2018interactive} showed that the image segmentation performance was improved by fine-tuning the CNN model to the test image through user scribble. 
In Jang et al.'s study~\cite{jang2019interactive}, the image segmentation performance was improved by creating an interaction map based on the distance from the user input and reflecting it in a deep learning model through a backpropagating refinement scheme. 
In the study conducted by Sardar et al.~\cite{sardar2020iris}, the performance of iris segmentation was improved by fine-tuning the U-net through user input.
Many studies have been conducted to improve the performance of deep learning through user interaction, but no studies have been conducted on the neuronal structure and mitochondria analysis.

\section{Design Rationale}
\label{sec:task}
The target user of our system is a neuroscientist studying the neuronal mitochondria morphology. 
We analyzed the tasks and challenges to achieve their research goals through close collaboration and interviews with two neuroscientists, and defined a design goal to address them.

\begin{figure}
  \centering
  \includegraphics[width=0.7\linewidth]{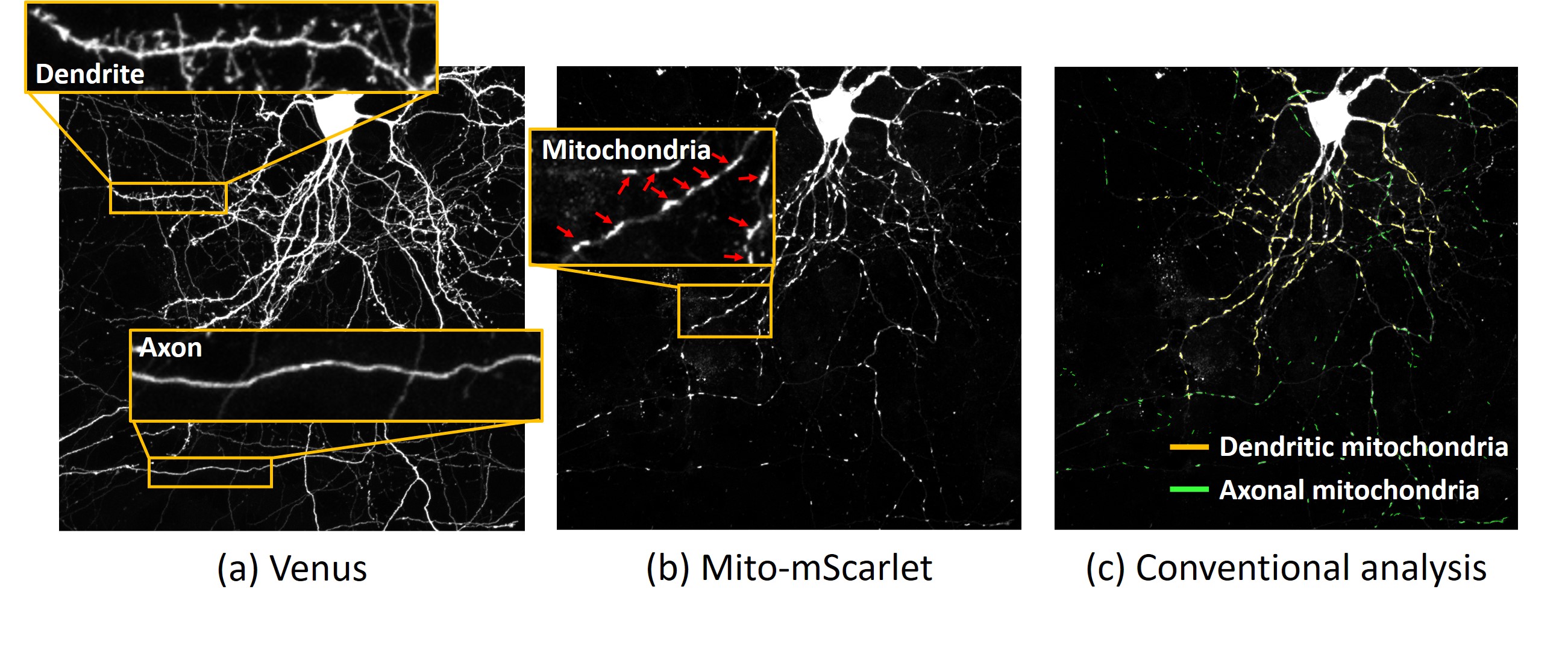}
  \caption{The target dataset of this study and conventional analysis. (a) Venus represents neuronal morphology such as dendrite and axon, and (b) Mito-mScarlet represents the mitochondria. (c) In conventional analysis, a user manually annotates dendritic mitochondria and axonal mitochondria using image processing tool.}
  \label{fig:dataset}
\end{figure}

\subsection{Research goal and target data}
\label{sec:dataset}

Our collaborators aimed to research the differences in mitochondria morphology between neuronal structures such as dendrites and axons. 
All datasets used in our experiment were acquired using the same protocol described in~\cite{lewis2018mff} as follows: 
the bicistronic plasmids expressing Venus (Fig.~\ref{fig:dataset}a) and mito-mScarlet (pCAG-Venus-T2A-mito-mScarlet) (Fig.~\ref{fig:dataset}b) were transfected into mouse cortical neurons using ex utero electroporation. 
Venus was used to visualize the neuronal morphology by filling the whole neuronal processes with yellow fluorescence, and mito-mScarlet, a mitochondria-targeting sequence-conjugated red fluorescence protein, was used to label mitochondria.


\subsection{Conventional analysis workflow}

In the conventional mitochondria analysis workflow, the experts manually annotated mitochondria located in axons and dendrites using commonly used image editing software, such as ImageJ~\cite{wiemerslage2016quantification,merrill2017measuring}. 
Once the manual annotation was performed, morphological features from annotated mitochondria were extracted and further analyzed using statistical-analysis software such as Microsoft Excel. 
Even though biologists are aware of recent deep learning algorithms, these algorithms are not yet actively employed in the current analysis workflow because of the steep learning curve for non-machine-learning experts.
The output of deep learning methods is hardly perfect and still requires time-consuming and laborious manual error correction, which also hinders the adaptation of automated algorithms in the conventional analysis workflow.


\subsection{Tasks and challenges}
We have analyzed the required tasks and challenges 
in the current analysis workflow. 
\begin{itemize}
    \item Structure segmentation: To analyze the morphology of mitochondria in neurons, the first step is to perform image segmentation to separate and extract individual neuronal structures. 
    %
    Even though several recent studies have proposed deep-learning-based neuronal structure segmentation methods~\cite{chan2021noiseloss}, applying a pre-trained model to new target data at the inference time requires additional model adaptation (or re-training), which is often conducted as a slow offline process. 
    Moreover, even though the deep learning model is well trained, there are always inference errors, which should be manually corrected by a human for accurate analysis.
    Leveraging deep learning in the analysis workflow requires domain-specific knowledge (e.g., programming, parameter tuning, etc.), which is an additional hurdle for biologists who are non-computer-science majors.


    \item Mitochondria detection:  Once individual neuronal structures are extracted, the mitochondria in each structure should be detected. 
    Neuronal mitochondria are typically thin and narrow, and the image quality can vary owing imaging/staining artifacts and noise. 
    Moreover, we deal with 2D images where neurons cross and overlap each other. 
    Therefore, similar to neuron structure segmentation, manual error correction is a necessary task in mitochondria detection. 
    
    
    \item Morphology analysis: Morphology analysis includes statistical analysis of the differences in mitochondria morphology between neuron structures (e.g., dendrites and axons), multiple neurons, or specific parts of neurons (e.g., proximal dendrite, one neurite, etc.). 
    In previous studies~\cite{lewis2018mff,fecher2019cell,kim2020altered}, task-specific analysis processes have been proposed, hindering the wider adaptation of the method to various analysis scenarios.
    %
    %
    Moreover, conventional analysis relies on existing stand-alone software tools, which require extra data management between the analysis tasks. 
    %
    %
    Owing to this bottleneck, conventional analysis requires a longer time for the entire analysis, which makes it difficult to perform large-scale analyses.
    
\end{itemize}

\subsection{Design goals}

Through the above task and challenge analysis, we defined the design goals of our system. We primarily aimed to enable users to utilize the latest deep learning technologies easily and effectively through visual guides and analytics systems, and to perform the necessary analysis flexibly and quickly to realize precise large-scale analysis. The detailed design goals are as follows:

\begin{itemize}
    \item Easy use of deep learning method: It should be possible to optimize a pre-trained deep learning model to the user's data presented at the inference time through an intuitive and easy method without a low-level process. 
    This allows users without specialized knowledge about deep learning to effectively use deep learning in their analysis workflow. 
    We aim to make this process possible with only intuitive interaction and obtain highly accurate results for data even with a different distribution from the training dataset of the deep learning model.
    
    \item Easy and fast proofreading: Deep learning method has the potential to contain errors owing to data noise, artifacts, etc. These errors must be proofread by the user prior to the analysis. 
    We aim to enable users to accurately correct errors that exist in the results through deep learning as much as possible and only with simple and intuitive interactions.
    
    \item Effective, diverse, and precise morphology analysis: Along with the results of deep learning, various analyses must be able to be performed precisely and effectively. 
    Users should be able to configure the desired analysis process easily and perform a large-scale statistical analysis. In addition, users should be able to receive immediate feedback on the generated analysis data (e.g., mitochondria subsets); so that the users can check whether each step is performed as intended, and the entire process should be easily managed.
\end{itemize}

\section{MitoVis}

Based on the task analysis, we developed MitoVis that can effectively perform all data processing and analysis tasks from microscopic images to precise large-scale statistical analysis. 
The workflow of MitoVis is as follows. 
%
\begin{enumerate}[label=\arabic*),itemsep=-0.5ex]

\item With microscopic images, initial structure segmentation and mitochondrial detection are performed using a pre-trained deep learning model. 
\item User verifies the result by exploration of the image and result label through various visualizations. 
\item With a visual guide, the user can annotate the error part of the result from deep learning. 
\item The deep learning model is fine-tuned in real time through the partially annotated error part, and the resulting model is applied to the entire image to derive a new result. 
\item 2-4 steps are repeated until the optimized result is derived. 
\item Finally, statistical analysis is performed through the finally obtained structure label and mitochondria objects with various visualization techniques.
\end{enumerate}

\begin{figure}
  \centering
  \includegraphics[width=0.7\linewidth]{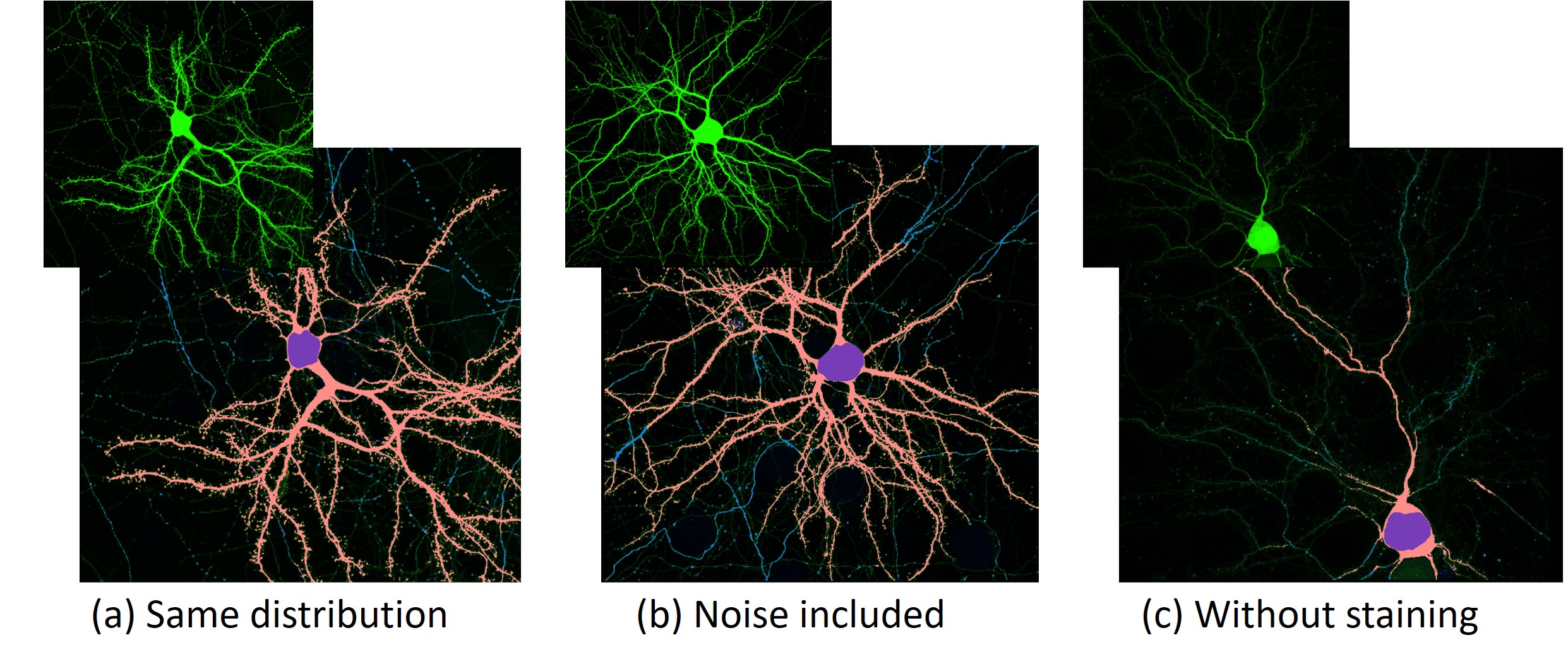}
  \caption{Possible errors in deep learning depending on the data. If noise is included in the data as in (b) or unseen data that is not included in the training data as in (c), many errors can be included.}
  \label{fig:error_example}
\end{figure}

\subsection{Initial structure segmentation and mitochondria detection}

When the user loads the raw image from the tool, the dendrite, axon, cell body label and mitochondria foreground label are predicted through a pre-trained deep learning module~\cite{chan2021noiseloss,fischer2020mitosegnet}. 
Each mitochondria object is detected by a connected component from the mitochondria foreground label. 
%
Since the deep learning model is trained to produce optimal results for the training data, there are cases in which errors are included in the result when the user applies it to his or her own data.
%
Figure~\ref{fig:error_example} shows the examples of applying the pre-trained deep learning model to real-world data. As the conditions or characteristics of the data are different from the training set, many errors are included. 
In particular, Figure~\ref{fig:error_example}a contains partial errors due to the noise of the image even though it is acquired under the same conditions as the training dataset. 
To perform precise analysis, these errors must be corrected. In the following sections, a framework for correcting these errors is introduced.

\begin{figure}
  \centering
  \includegraphics[width=0.7\linewidth]{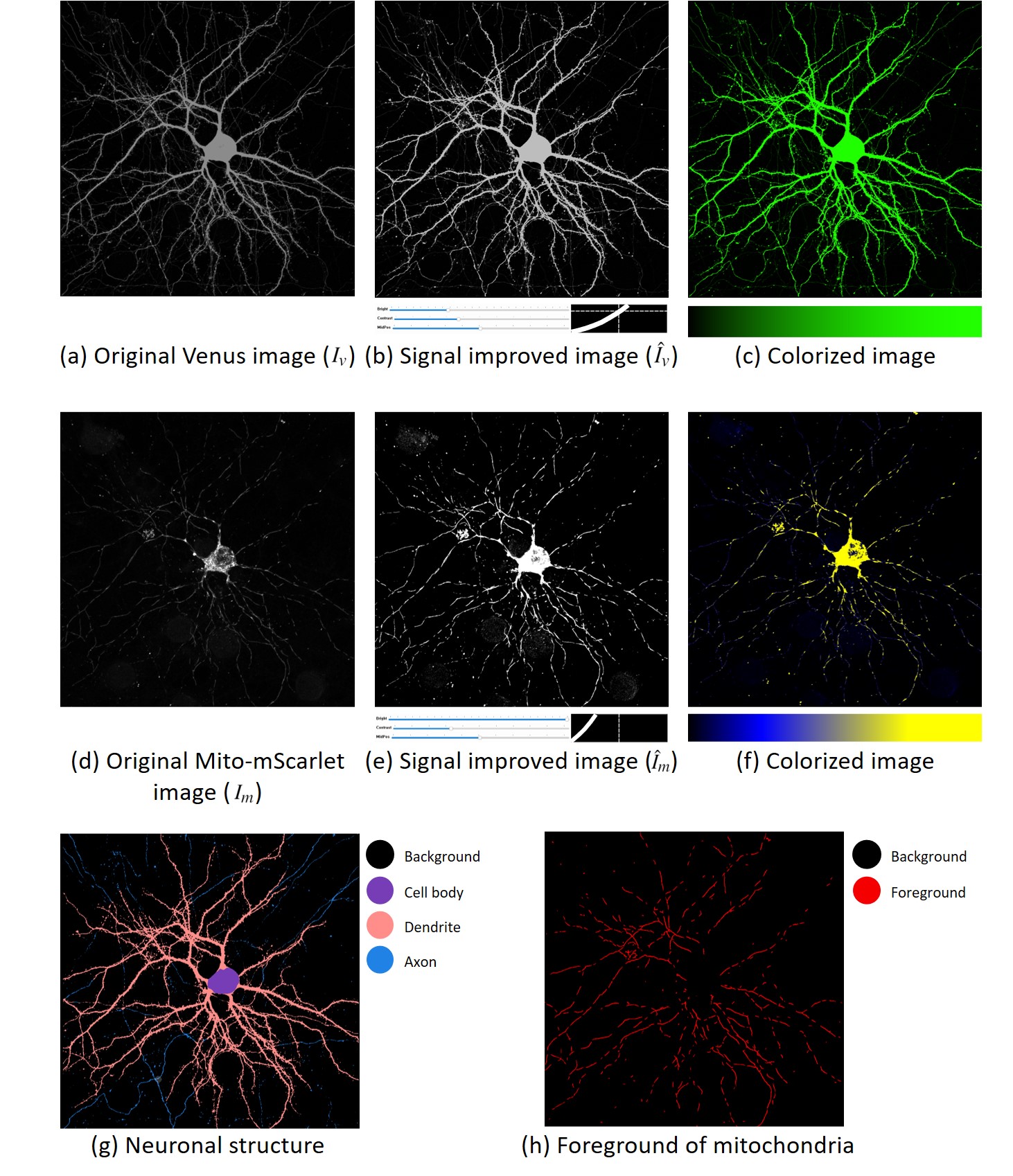}
  \caption{Visualizations for image data exploration. The four colorized images (c, f, g, and h) are blended with user-specified blending factors to generate a final image visualization.}
  \label{fig:image}
\end{figure}

\subsection{Image data exploration}

The image data utilized by MitoVis is the Venus $(I_v)$ and the mito-mScarlet $(I_m)$ as inputs, and a structure segmentation label $(L_s)$ and a mitochondria object label $(L_m)$ obtained through deep learning. 
The signal of $I_v$ and $I_m$ is 
enhanced by the signal improving function 
with user parameters as follows: 
\begin{equation} \label{eq1}
\begin{split} 
\hat{I}_v=\frac{2b_v}{1+e^{-60c_v(I_v-t_v)}}
\\ \hat{I}_m=\frac{2b_m}{1+e^{-60c_m(I_m-t_m)}}
\end{split}
\end{equation}
where $\hat{I}_v$ and $\hat{I}_m$ are signal improved image, $b_v$ and $b_m$ are brightness, $c_v$ and $c_m$ are contrast, and $t_v$ and $t_m$ are translate factor. All variable is between 0 and 1. 
Examples of the signal improved images are shown in Fig.~\ref{fig:image}b and e.

\begin{figure}
  \centering
  \includegraphics[width=0.7\linewidth]{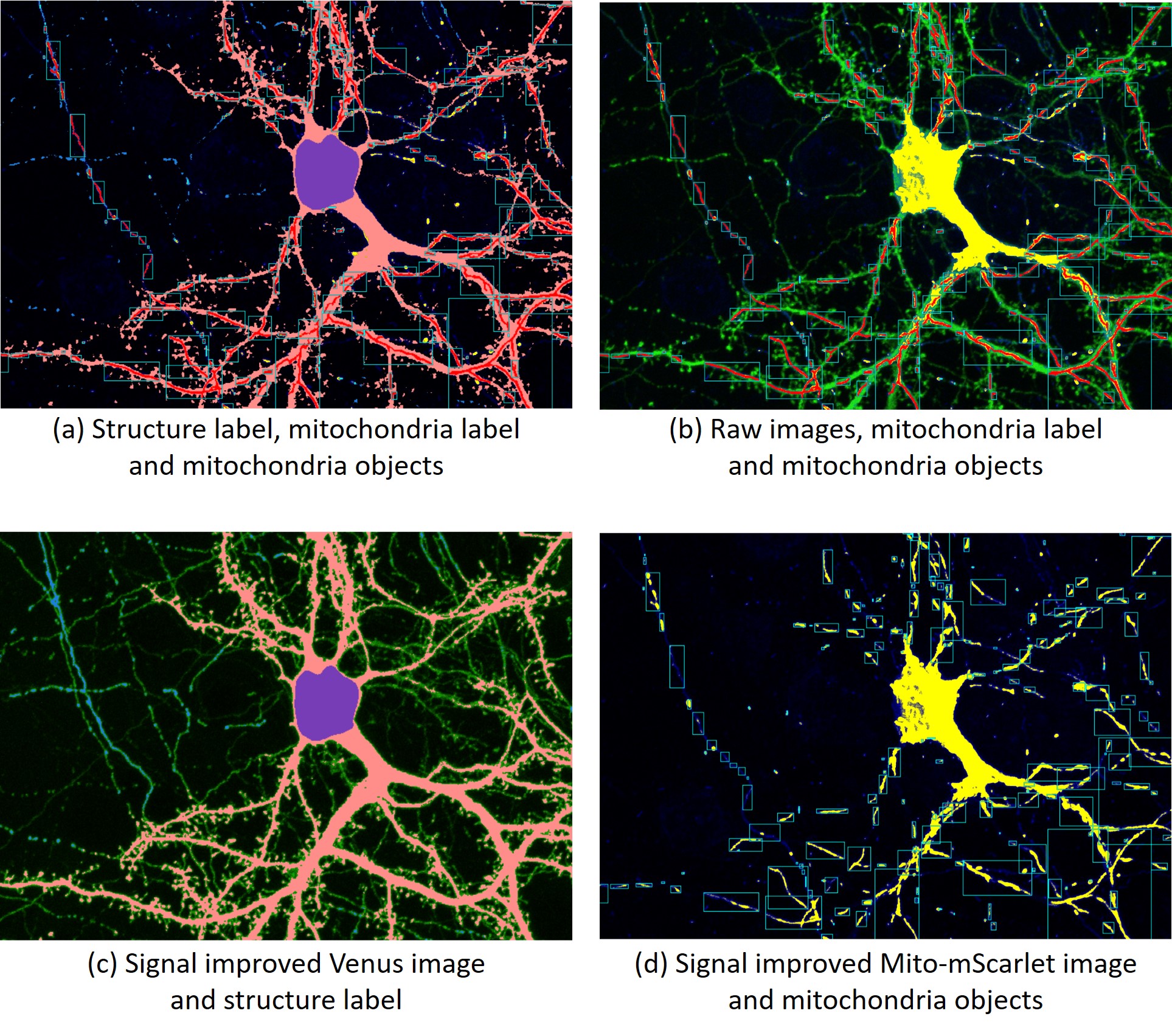}
  \caption{Examples of task-specific visualization setting. 
  Users can generate an effective visualization for a specific task by adjusting visualization parameters, such as brightness, contrast, and blending factors.}
  \label{fig:visualization_example}
\end{figure}

$\hat{I}_v$,$\hat{I}_m$,$L_s$, and $L_m$ are converted to color values using each color-map (Fig.~\ref{fig:image}c,f,g,h), and these are blended by user-defined transparency to create the final image. 
Depending on the preference or task, the user adjusts the transfer function and transparency of each image and explores the image data. Figure~\ref{fig:visualization_example} shows an example of a user-defined visualization according to tasks. Details about the visualizations and tasks are discussed in Section~\ref{sec:case_study}.

\subsection{Visual guiding and error correction}

Since the results obtained through deep learning are prone to noise, a proofreading process is necessary for practical use in analysis. 
Because our target data contains complex neuronal structures and numerous mitochondria, it is quite difficult and inefficient for users to find all errors with the naked eye. 
Therefore, we propose novel visual guides suitable for neuronal structures and mitochondria objects.
%
These visual guides provide spatial clues to the users to quickly identify the locations of errors in the structure and mitochondria labels, which can be corrected via user interactions.


\textbf{Structure label correction: }The process to correct the structure label is as follows; 

\begin{figure}
  \centering
  \includegraphics[width=0.7\linewidth]{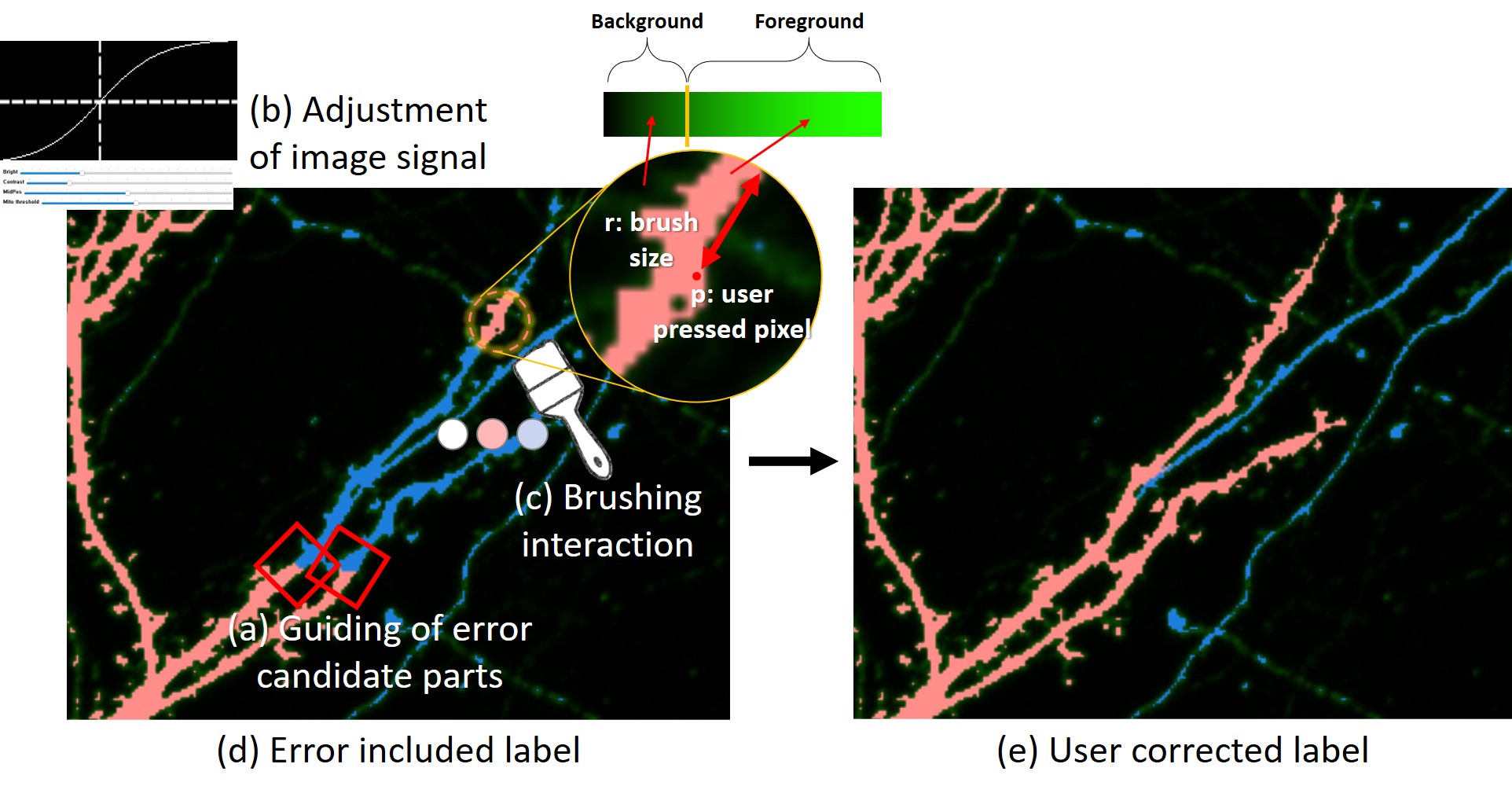}
    \caption{Structure label correction. (a) Error candidates are highlighted with visual guiding. (b) Venus image signal can be enhanced to clearly recognize the errors. (c) User can correct the error parts with a brushing interaction. }
  \label{fig:label_correction_interaction}
\end{figure}

1) Find the error part through visual guiding for the error candidate (Fig.~\ref{fig:label_correction_interaction}a). We hypothesized that an error is likely to occur in a place where multiple labels are mixed through the characteristic of a neuron in which the structure is continuous. Therefore, visual guiding for error candidates is given as a red box on the mixed label; 

2) For too strong or weak signal parts, adjust parameters to correct the image signal to an appropriate level (Fig.~\ref{fig:label_correction_interaction}b); 

3) Set the brush type as the target label to be corrected (Fig.~\ref{fig:label_correction_interaction}c); 

4) By dragging the part to be corrected on the structure label image, the pixels belonging to the threshold among the pixels in the brush area are corrected with the corresponding label (Fig.~\ref{fig:label_correction_interaction}e). The pixels ($B$) corrected through the brushing interaction are determined by the equations below.

\begin{equation} \label{eq6}
\begin{split} 
B(u) = 
\begin{dcases}
\{ i\in [1,m]\ \mid\ d(i,u) < r,\ i \in CC_v(u,\ <\sigma_s)\},\\  \ \ \ \ \ \ \ \ \ \ \ \ \ \ \ \ \ \ \ \ \ \ \ \ \ \ \ \ \ \ \ \ \ \ \ \ \ \ \ \ \ \ \ \ \ \ \ \textrm{if brush type is background.} \\
\{ i\in [1,m]\ \mid\ d(i,u) < r,\ i \in CC_v(u,\ \geq\sigma_s)\},\ \textrm{otherwise.}
\end{dcases}
\end{split}
\end{equation}


$m$ is a size of vectorized image ($m = width\times height$), $u$ is a index of pixel which is clicked by the user, $d(i,u)$ is the euclidean distance function between two indexed pixels, $r$ is a brush size, $\sigma_s$ is a threshold value to divide foreground and background for structure (yellow line on the color-map of Fig.~\ref{fig:label_correction_interaction}c), $CC_v(u,<\sigma_s)$ is a connected component function that results in a connected component consisting of values smaller than $\sigma_s$ from $u$ in $\hat{I}_v$, and $CC_v(u,\geq \sigma_s)$ is a function that results in a connected component consisting of values equal or higher than $\sigma_s$ from $u$ in $\hat{I}_v$. This approach minimizes the situation in which unintended parts are modified.

\begin{figure}
  \centering
  \includegraphics[width=0.7\linewidth]{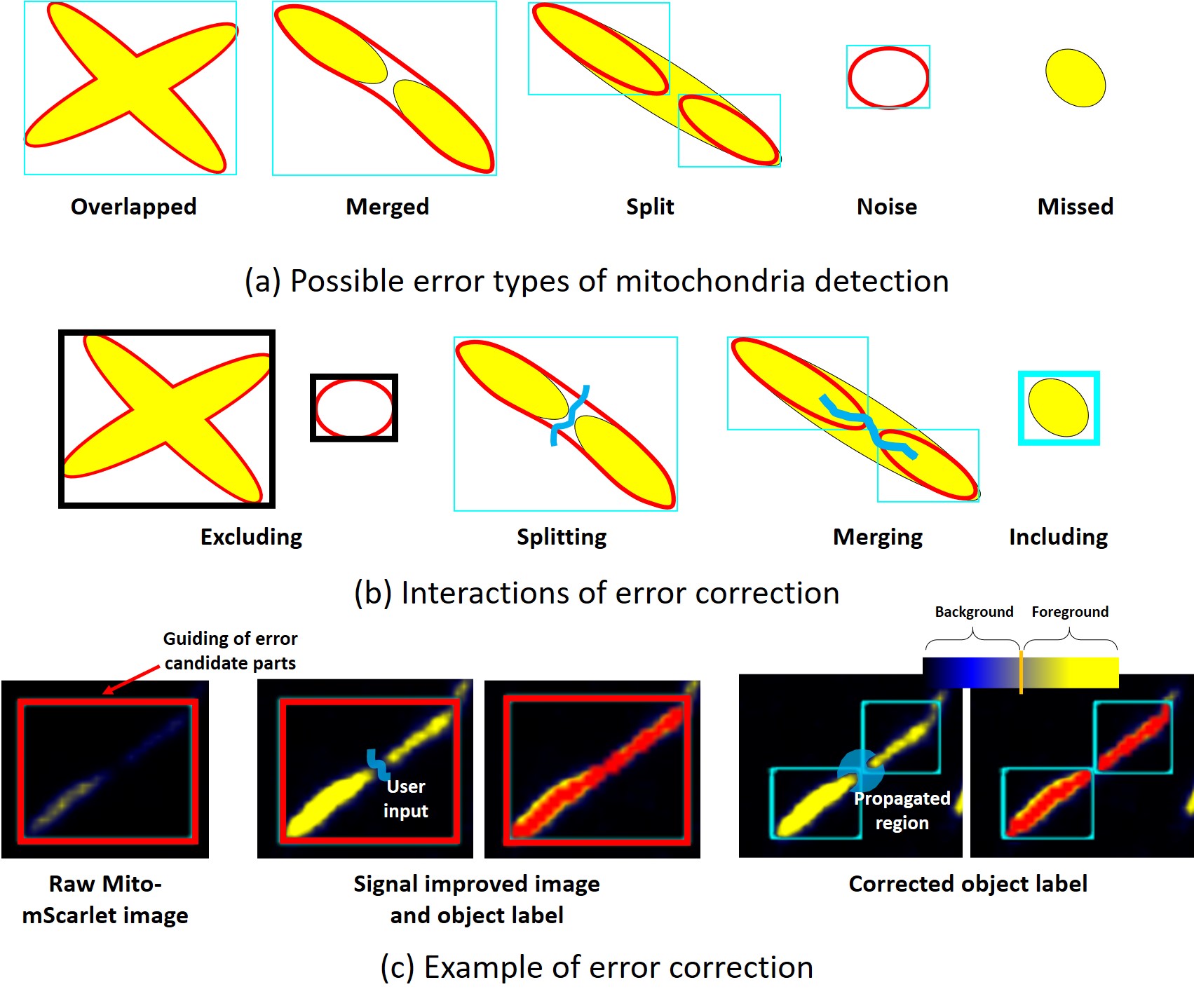}
  \caption{Mitochondria objects correction. (a) The possible error can be defined as five types. (b) Four interactions are designed to effectively correct the five error types. (c) A splitting interaction can effectively correct a merged error type.}
  \label{fig:mito_correction_interaction}
\end{figure}

\textbf{Mitochondria label correction: }We defined 5 cases in which mitochondria objects are incorrectly detected; overlapped object error occurring in the overlapped part of the structure signal due to complex data; merged object error when two objects are detected as one; split object error in which one object is detected as two objects; noise error that non-mitochondria is detected; missing error that failed to detect mitochondria (Fig.~\ref{fig:mito_correction_interaction}a). The process to correct these errors is as follows. 

1) Find the error part through visual guiding for the error candidate (Fig. ~\ref{fig:mito_correction_interaction}c). We hypothesized that the place where the signal is high in the mito-mScarlet image is likely to be the foreground of mitochondria. Therefore, if there are huge region with low signal in each mitochondria object area, there is a high possibility of the merged error or the noise error. If there are many high signals around each mitochondria object area, it is likely the split error. If there is a high signal in the background area, there is a possibility of the missed error. Through these observations, we defined the error probability for detected objects ($E_o$) and the error probability for the background region ($E_b$).

\begin{equation} \label{eq8}
\begin{split} 
N(k)=\{i \in & [1,m]\ \mid\ i \in CC_m(j,\ \geq\sigma_m)\ where\ j \in O^{(k)} \}
\\ &E_o^{(k)} = 1-Dice(O^{(k)},N)
\\
&E_b^{(l)} = 
\begin{dcases}
0,\ if\ O_b^{(l)} \cap L_m \ne \emptyset 
\\ 1,\ otherwise
\end{dcases}
\end{split}
\end{equation}

$O^{(k)}$ is the k-th detected mitochondria object, $Dice$ is the dice coefficient function, $\sigma_m$ is a threshold value to divide foreground and background for mitochondria (yellow line on the color-map of Fig.~\ref{fig:mito_correction_interaction}c), and $O_b^{(l)}$ is the $l$-th object obtained through the connected component based on $\sigma_m$ in $\hat{I}_m$.
If $E_o^{(k)}$ or $E_b^{(l)}$ is greater than the error threshold ($\sigma_e$) specified by the user, $O^{(k)}$ or $O_b^{(l)}$ is highlighted in a red box as error candidates for visual guiding.

2) To correct the error parts, four interactions (excluding, splitting, merging, and including) were designed (Fig.~\ref{fig:mito_correction_interaction}b). The user can select overlapped objects or noise objects and correct them as the background through the excluding interaction.
Splitting interaction divides one object into two objects based on the drawn line ($u_l$) from the user. At this time, in the $\hat{I}_m$, the propagated region on the mitochondria label through the connected component from $u_l$ is corrected to the background (Fig.~\ref{fig:mito_correction_interaction}c). The propagated region ($P_b$) is as follows.

\begin{equation} \label{eq10}
\begin{split} 
P_b(u_l) = \{i \in & [1,m]\ \mid\ i \in CC_m(j,\ <\sigma_m)\ where\ j \in u_l,\ d(i,j)<5\}
\end{split}
\end{equation}

Merging and including interaction, contrary to excluding interaction, creates a propagated region ($P_f$) by performing a connected component from the user's drawing input, $CC_m(u,<\sigma_s)$ is a connected component function that results in a connected component consisting of values smaller than $\sigma_m$ from $u$ in $\hat{I}_m$, $CC_m(u,\geq \sigma_s)$ is a function that results in a connected component consisting of values equal or higher than $\sigma_m$ from $u$ in $\hat{I}_m$, and modifies it to the foreground. $P_f$ is as follow

\begin{equation} \label{eq11}
\begin{split} 
P_f(u_l) = \{i \in & [1,m]\ \mid\ i \in CC_m(j,\ \geq\sigma_m)\ where\ j \in u_l,\ d(i,j)<5\}
\end{split}
\end{equation}

These interactions are intuitive and make it possible to perform effective mitochondria object correction with few inputs.

\begin{figure*}
  \centering
  \includegraphics[width=\linewidth]{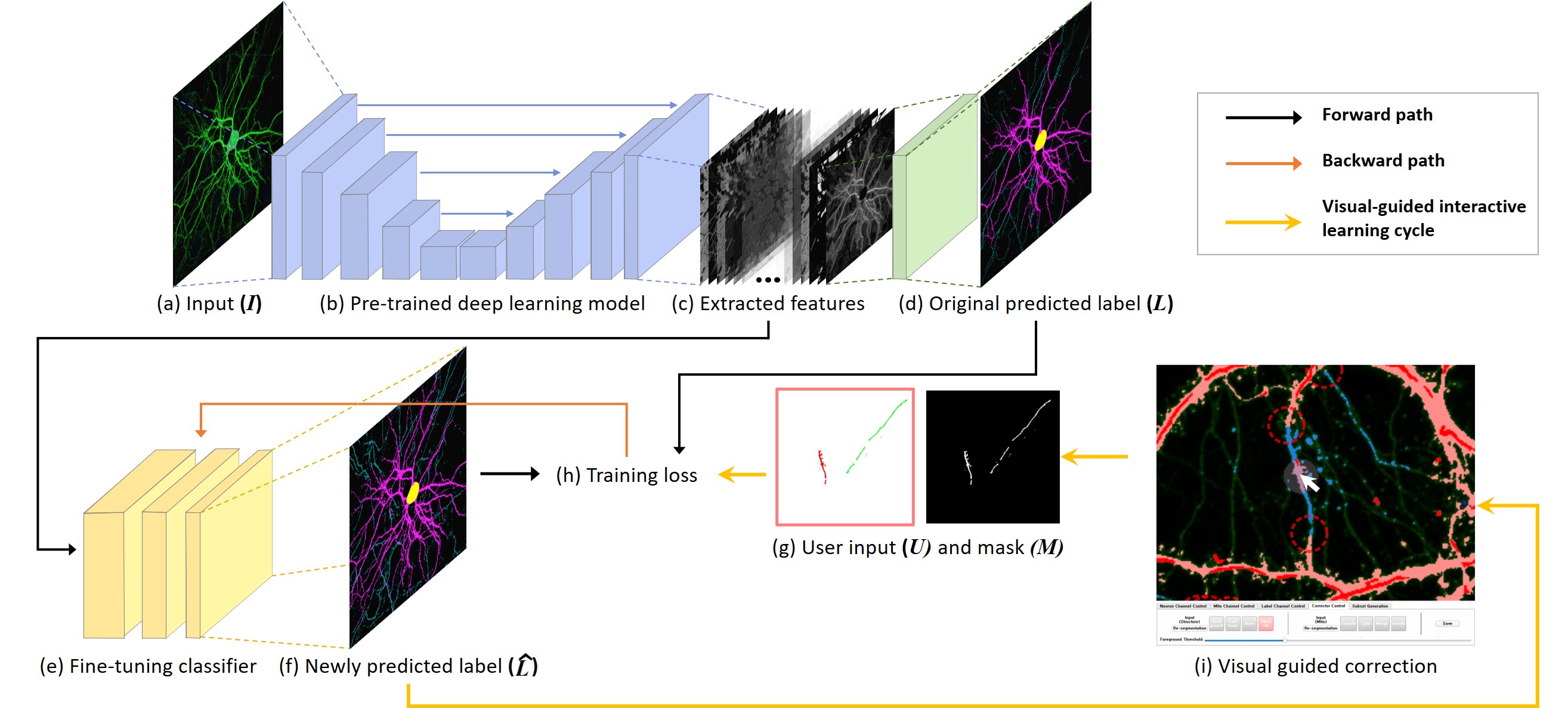}
  \caption{Visually-guided interactive learning framework. A input data (a), Venus or mito-mScarlet, is segmented by pre-trained deep learning model (b) for generating initial predicted label (d). Features obtained by a feature extraction part of the pre-trained deep learning model are used as a one input for fine-tuning classifier (e). The fine-tuning classifier can be trained with the features and user input (g) by correction in visual interface (i). A newly predicted label (f) is passed to the visual interface for updating the visualizations.}
  \label{fig:transfer_learning}
\end{figure*}

\subsection{Interactive learning}

It is inefficient to correct all errors included in the image through only user interaction. The more data to be analyzed, the greater the inefficiency becomes. If the distribution of the analysis data is different from the distribution of the training set of the pre-trained deep learning model and contains many errors in the results, it is almost impossible to correct all errors manually.
In order to solve this problem and obtain precise results with only a few interactions, we designed a method of fine-tuning a deep learning model interactively through user input.

\textbf{Challenges: }For that, a model should be able to be trained with a small amount of user input, and training and deployment should be completed within a reasonable time to modify user input with fine-tuning results interactively. In addition, the distribution of the image should be reflected in the model through fine-tuning and partial errors existing in the current data should be corrected at the same time.

\textbf{Model design: }To handle small inputs, we utilized an approach to training by fixing the feature extraction part of the pre-trained model and adding a new classifier, which has been widely used in recent few shot learning~\cite{sun2019meta}. Pixel-wise high-dimensional features are obtained through feature extraction parts from pre-trained models (Fig.~\ref{fig:transfer_learning}c). 
The new classifier model (Fig.~\ref{fig:transfer_learning}e) classifies these features. With the training loss (Fig.~\ref{fig:transfer_learning}h) by user input (Fig.~\ref{fig:transfer_learning}g), the classifier is trained. 
Since the classifier must be able to train within a reasonable time, it should not use a complicated network structure. 
In the mean time, 
a simple (weak) network model is not suitable either for obtaining sufficient accuracy.
%
As an appropriate model that satisfies these conditions, we use a simple convolutional neuronal network with three convolution layers.

\textbf{Training loss: }For training this model, the interaction loss ($Loss_u$) for focusing user input and the original loss ($Loss_o$) for maintaining the trend of the previous label were applied. $Loss_u$, $Loss_o$ and total loss ($Loss_t$) are defined as follows

\begin{equation} \label{eq12}
\begin{split} 
Loss_u= \|M\odot(U – \hat{L})\|_2 \\
Loss_o= \|(1-M)\odot(L – \hat{L})\|_2 \\
Loss_t=f Loss_u + Loss_o
\end{split}
\end{equation}

$ \odot $ is element-wise product, $U$ is the user input generated by error correction interaction, $M$ is a mask which converts user input as foreground, $L$ is a previous label from the pre-trained network, $\hat{L}$ is a new label, and f is a focusing factor that indicates how much focusing to the user input.


\textbf{Visually-guided interactive learning cycle: }The above label correction process creates a correction cycle as shown in Fig.~\ref{fig:transfer_learning} yellow lines.
With the help from visual guides, the user can interactively make annotations to correct errors in the structure and mitochondria labels, which are then used to re-train (fine-tune) the classifiers.
%
%
The fine-tuned deep learning models provide better structure labels and detected mitochondria objects, which are used to update the visual guides for the next iteration. 
The user can iterate this process until a satisfactory result is generated. 
%


\subsection{Mitochondria morphology analysis}

\begin{figure}
  \centering
  \includegraphics[width=0.7\linewidth]{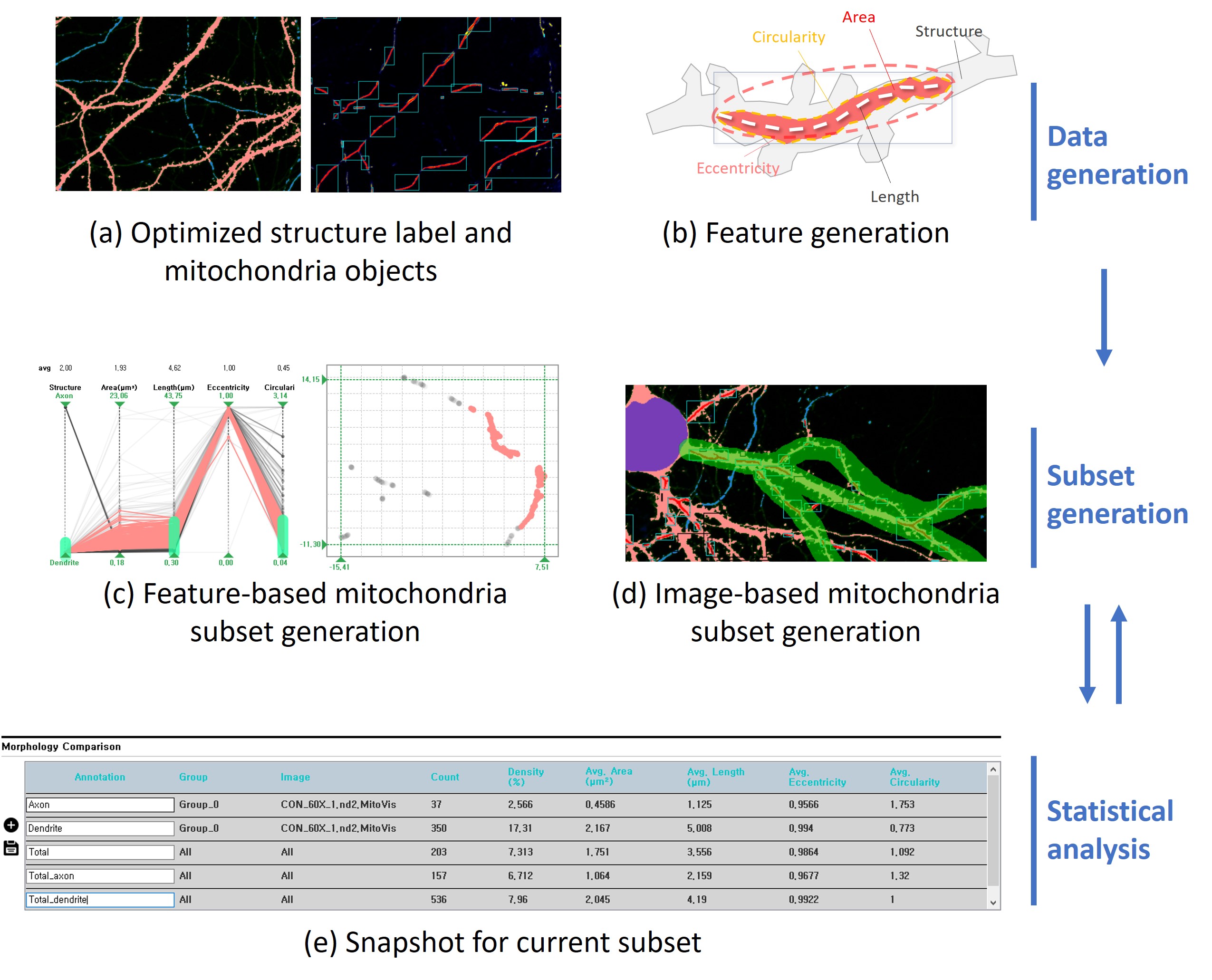}
  \caption{Mitochondria morphology analysis. (b) Morphological features are extracted from detected mitochondria objects (a), and users can generate diverse subset using feature-based (c) and image-based (d) subset generation functions. (a) Morphological information of the subsets can be recorded in a morphology comparison panel.}
  \label{fig:analysis}
\end{figure}

The mitochondria morphology analysis process includes that feature generation from the structure labels and mitochondria objects obtained through the previous process (Fig.~\ref{fig:analysis}b), generating mitochondria subsets  
(Fig.~\ref{fig:analysis}c,d), and performing morphology comparison between various subsets through statistical snapshot recording function (Fig.~\ref{fig:analysis}e).

\textbf{Feature generation:} In this study, a total of five features are used for analysis, such as four morphological features (area, circularity, eccentricity, and length) and one feature of a structure label (Fig.~\ref{fig:analysis}b).

\textbf{Subset generation:} Users can create subsets in feature space through parallel coordinate plot (PCP) and dimensional reduction plot (DRP). In each axis of PCP, objects belonging to a specific range can be selected, and in DRP, objects belonging to a specific area on the plot can be selected. As a method for generating DRP, a dimensional reduction method such as PCA~\cite{bro2014principal} and UMAP~\cite{mcinnes2018umap} can be applied, or two features can be selected to create the plot. 
In addition, the user can create a subset in the image space by brushing a specific area in the image viewer. 
This can be useful when the user wants to perform an analysis on a specific area such as proximal dendrite or distal dendrite. 
By combining these subset generation methods, users can easily generate desired mitochondria subsets.

\textbf{Morphology comparison:} Morphology comparison can be performed by recording specific subsets as a statistical snapshot. For example, the user can create and record the mitochondria subset located in the axon and the mitochondria subset located in the dendrite, respectively, and analyze the difference in mitochondria morphology located in the dendrite and the axon. In the snapshot, user comments, information of the dataset, count, density, average area, average length, average eccentricity, and average circularity of mitochondria objects in the current subset are recorded.
The density is obtained as the ratio of the mitochondria object area and the structure area where the mitochondria object is located.

\subsection{Visual interface}

\begin{figure*}
  \centering
  \includegraphics[width=\linewidth]{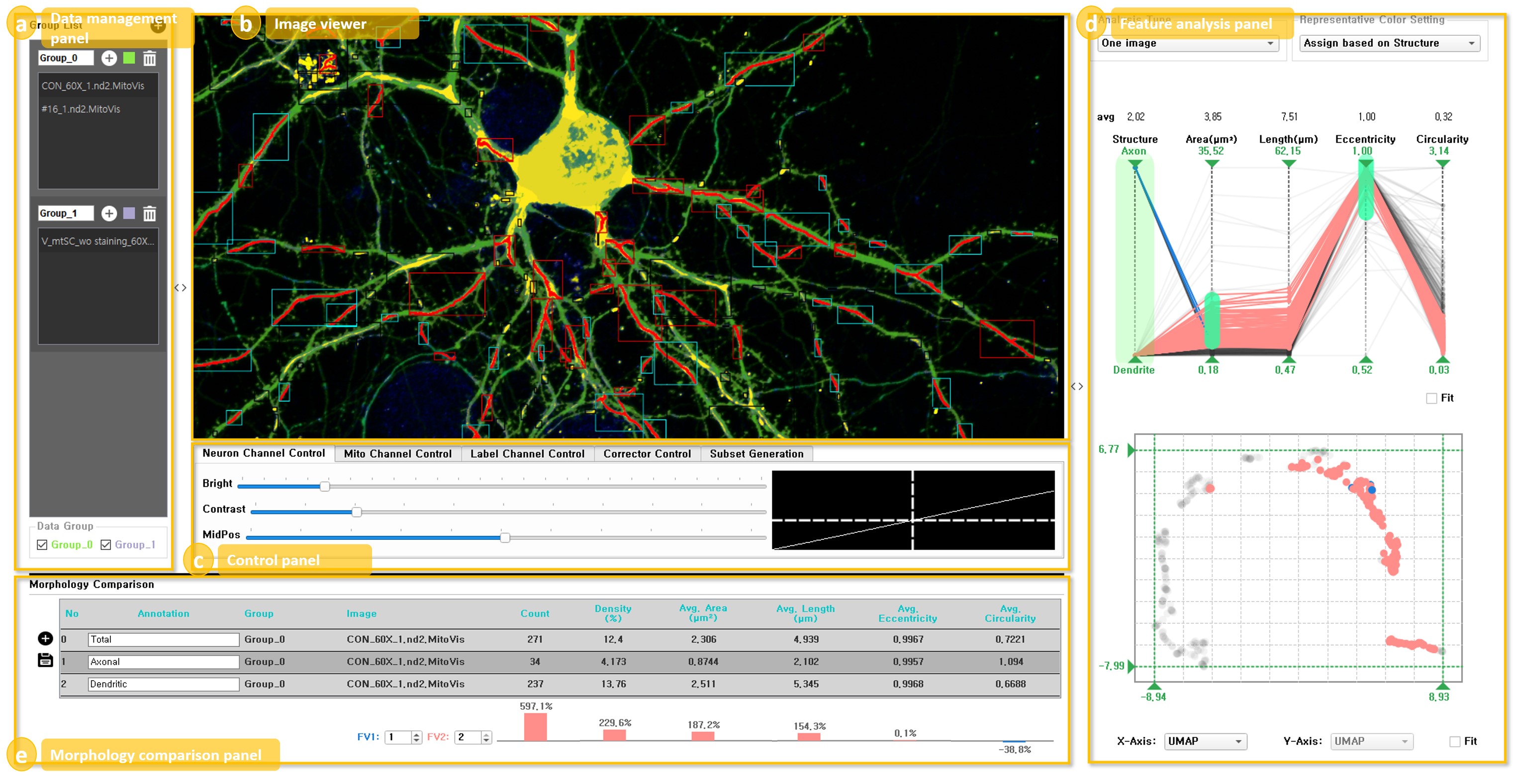}
  \caption{Visual interface of MitoVis. (a) Dataset can be loaded and managed in the data management panel. (b) In the image viewer, image visualizations are rendered and correction interactions are performed. (c) All parameters for visualizations and options for the correction can be adjusted in the control panel. (d) Morphological features of mitochondria can be explored and filtered in the feature analysis panel. (e) Morphological information can be recorded and compared in the morphology comparison panel.}
  \label{fig:visual_interface}
\end{figure*}

The user can perform all of the described methods in a single visual interface (Fig.~\ref{fig:visual_interface}).

\textbf{Data management panel:} In this panel, generated datasets can be managed by group, and the group to be used for analysis can be selected. Users can classify groups according to data analysis cases, image acquisition time, etc. Since each group can be analyzed as a single dataset, large-scale analysis is possible.

\textbf{Image viewer:} Here, the user can check all results of structure segmentation and mitochondria detection. In addition, image data exploration (Sec.4.2) and error correction (Sec.4.3) can be performed.

\textbf{Control panel:} The visualization displayed in the image viewer is created by controlling the image signal through various parameter adjustments and controlling the transparency of each visualization. In addition, user interaction can be performed by selecting the correction options or image based subset generation options.

\textbf{Feature analysis panel:} Data exploration and mitochondria subset generation can be performed based on features. In addition, it is possible to select whether to conduct data analysis in one image or based on groups.

\textbf{Morphology comparison panel:} Statistical information of morphology can be recorded for various mitochondria subsets generated through the above panels. Users can record this at any time during the analysis. Through this recorded information, morphology between various subsets can be compared.

\subsection{Implementation}

We used Park et al.~\cite{chan2021noiseloss} as a baseline pre-trained deep learning model with additional fine-tuning layers for structure segmentation, which is written in Python using PyTorch~\cite{paszke2019pytorch}.
%
%
%
As for the deep learning model for mitochondria segmentation, the model introduced in Fischer et al.~\cite{fischer2020mitosegnet} was re-implemented using PyTorch, and the hyperparameters were modified to fit to our data. 
This model was pre-trained using Venus and mito-mScarlet datasets. 
Considering that the general user's analysis environment uses a lightweight machine without a fast graphics card (e.g., laptop), all of our deep learning models for inference are designed to work on the CPU. 
In addition, the time required for fine-tuning was limited to a maximum of one minute (determined based on the feedback from the users of our system) for interactive analysis.
%
%
The visual interface was developed in C++ using QT and OpenGL libraries.

\begin{figure}
  \centering
  \includegraphics[width=0.7\linewidth]{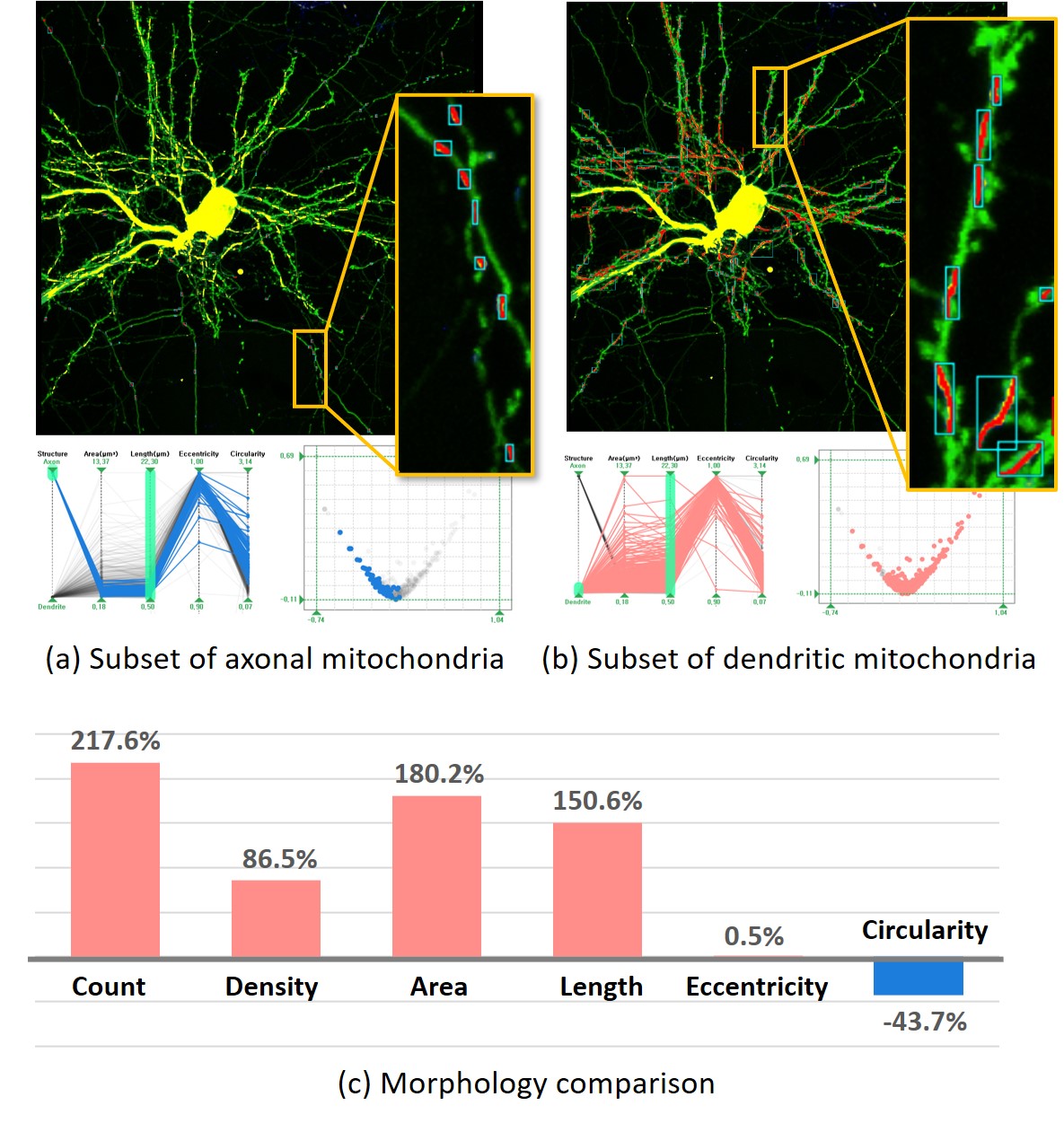}
  \caption{Morphology analysis result for axonal and dendritic mitochondria. Image visualization, PCP, and DRP by PCA for subset of axonal mitochondria (a) and subset of dendritic mitochondria (b). The difference of morphology between the two subset is calculated as $(fv_1-fv_2)\div fv_1 \times 100\%$, $fv_1$ is feature value of the subset of axonal mitochondria, and $fv_2$ is feature value of the subset of dendritic mitochondria.}
  \label{fig:case_study}
\end{figure}

\section{Case study and expert feedback}
\label{sec:case_study}

With a domain expert, we conducted a case study using real-world datasets to describe how MitoVis can utilize deep learning effectively, and how interaction and visualization facilitate the analysis.

\subsection{Experiment design}

\textbf{Participant: }The participant of this experiment is a neuroscientist who has been performing neuronal mitochondria analysis and only knows the overall concept of deep learning. The expert mentioned that deep learning has hardly been used in previous research because the part where deep learning can be used had not been defined, and there is no system that can effectively utilize the deep learning technology.

\textbf{Dataset: }In this experiment, we used three datasets of Venus and Mito-mScarlet, each 1024$\times$1024 in size. 
To acquire these images, cells were fixed at 17 days in vitro and stained to amplify the fluorescence signal. After staining, the samples were imaged using a Nikon A1R confocal microscope with a 60x oil lens. These images were not included in the pre-training data of the deep learning models.

\textbf{Task: }
With MitoVis, structure segmentation and mitochondria detection were performed on each dataset. 
Morphological features were obtained from the mitochondria located in axons and dendrites, and a comparative analysis of the morphology was performed. 
We compared the task-completion time and analysis accuracy with those of the conventional analysis (Fig.~\ref{fig:dataset}c). 
%
For the accuracy comparison, the average mitochondria lengths calculated from the manual annotation of the expert is used as a ground truth. 
%
MitoVis can also analyze other morphological features such as density, area, eccentricity, and circularity. %
However, only the length can be measured in the conventional analysis, so accuracy comparison was performed only for the length.
%

\textbf{Procedure: }
Before the actual user study started, we allowed the participant a warm-up time to familiarize themselves with the system. 
Subsequently, the above task was performed on the three images. 
The task-execution time was measured from the moment when each dataset was loaded until the morphological features of the axonal mitochondria and dendritic mitochondria were recorded.
After the experiment was completed, we received feedback through an interview about the usability and performance of MitoVis, the possibility in the research field, and the direction of development.

\textbf{Experiment environments: }This experiment was conducted on a laptop equipped with Intel i5-1035G7 CPU (1.20GHz) and 8GB RAM in the same environment as the user's usual analysis.

\renewcommand{\arraystretch}{1.2}
\begin{table}[]
\centering
\caption{Performance comparison between conventional analysis and MitoVis. Time is measured in minutes and accuracy is calculated as $(1-|len_c - len_m|\div len_c) \times 100\%$, $len_c$ is average mitochondria length by conventional method (manual measurement) and $len_m$ is average mitochondria length by MitoVis.}
\label{Tab:result}
\begin{tabular}{lccccc}
\hline
          & Conventional &  & \multicolumn{2}{c}{MitoVis} & \multirow{2}{*}{Speed up} \\ \cline{2-2} \cline{4-5}
          & Time &  & Time & Accuracy &                           \\ \hline
dataset 1 & 100 &  & 18 & 92.3 & 5.6\\ 
dataset 2 & 153 &  & 10 & 66.5 & 15.3\\ 
dataset 3 & 170 &  & 12 & 86.1 & 14.2\\ \hline
\end{tabular}
\end{table}

\subsection{Axonal and dendritic mitochondria morphology comparison}

\textbf{Initial structure segmentation and mitochondria detection: }An analysis group was created in the MitoVis data management panel, and Venus and mito-mScarlet images were loaded into the group. A structure label and mitochondria foreground label were automatically generated from the pre-trained deep learning model. In addition, mitochondria objects were created from the mitochondria foreground label through connected components, and morphological features for each object were created. The average time required for this step was 43 seconds.

\textbf{Error correction: }The expert adjusts the image signal improvement parameters and blending parameters of the control panel to find errors. 
%
For the structure label error, the expert searched by adjusting the transparency of the structure label in the visualization shown in Fig.~\ref{fig:visualization_example}c, which mainly blended the signal-improved Venus image and the structure label. 
%
For errors in mitochondria objects, visualization such as that shown in Fig.~\ref{fig:visualization_example}b or d was mainly used. 
Through the visualization in Fig.~\ref{fig:visualization_example}b, it was easy to find the overlapping error that appears where the neuron structure is entangled or the noise error that exists outside the neuron structure. 
Fig.~\ref{fig:visualization_example}d presents a visualization wherein only the bounding box of the mitochondria object is activated with the signal improved mito-mScarlet image, making it easy to find the merged error. 
In addition, the expert could find the error more easily by first exploring the highlighted area by following the visual guides. 
The errors were corrected by structure label correction and mitochondria object correction interaction.

\textbf{Interactive learning: }After completing the error correction partially, if the user clicks the re-segmentation button on the control panel, the fine-tuning classifier is trained for approximately one minute, and a new label for the entire image is created through the trained fine-tuning classifier. 
However, in the data of this experiment, there were only a small number of errors in the initial predicted labels. 
Therefore, unlike our expectation, all the errors were manually corrected using interaction and visual guidance without leveraging interactive learning. 
We discuss more about the use case of interactive learning in Section~\ref{sec:discussion} 'Interactive learning approach' paragraph.


\textbf{Mitochondria subset generation: }With the clean structure label and mitochondria objects created in the previous step, the analysis was performed. First, only dendrites were activated among the structure features through the filtering function in the PCP on the feature analysis panel. 
In addition, to exclude tiny objects, only the objects whose length is greater than 0.5$\mu$m are activated. Thus, the expert could create a dendritic mitochondria subset, and the morphological information of the subset was recorded in the morphology comparison panel through the snapshot function. Similarly, a axonal mitochondria subset was generated and recorded in the morphology comparison panel.

\textbf{Morphology comparison: }Figure~\ref{fig:case_study} presents the results obtained through this process. Through image data exploration, it was confirmed that axonal mitochondria were shorter and more fragmented, compared to dendritic mitochondria. As for the morphological features, the dendritic mitochondria are much larger in terms of the object count, density, area, and length. The circularity is larger in axonal mitochondria, indicating that the axonal mitochondria are short and rounded.

\subsection{Performance analysis}

Table~\ref{Tab:result} shows the performances of the conventional analysis method and MitoVis. The conventional analysis methods took 100, 153, and 170 for each dataset, whereas MitoVis took 18, 10, and 12 min, leading to an increase of up to 15 times the analysis speed. As for the significant improvement in analysis speed, the expert reported that the time-consuming pre-processing task can be significantly shortened through deep learning, and the visual guide makes it possible to find the error part more quickly. 

The analysis accuracy of MitoVis was 92.3\%, 66.5\%, and 86.1\% for each dataset. The average length of axonal mitochondria was 1.28, 1.47, and 1.34$\mu$m in the conventional analysis, and 2.17, 1.84, and 1.68$\mu$m in MitoVis. The average length of dendritic mitochondria was 3.65, 3.85, and 3.99$\mu$m in the conventional analysis, and 3.89, 4.72, and 4.21$\mu$m in MitoVis. 
About this result, the expert commented that MitoVis tended to take slightly longer to detect the mitochondria (1$\sim$4 pixels, 1 pixel is 0.21$\mu$m), but it is not significant 
and the mitochondria shapes are generally detected well.
%

Both the conventional analysis and MitoVis confirmed that the dendritic mitochondria were much longer than the axonal mitochondria, showing statistically significant difference (p-value by t-test are 0.0001 and 0.0005, respectively). 
The difference between the average length of axonal and dendritic mitochondria was 2.47$\mu$m in the manual method and 2.38$\mu$m in MitoVis.

\subsection{Expert feedback}

The expert reported that the biggest advantage of MitoVis is that it can quickly analyze images for various subsets and groups, which is 
%
useful for various data screening tasks in mitochondria research. 
%
%
Other advantages of MitoVis are that the system is very easy to learn and use because the interaction and all functions are intuitively designed; it is possible to utilize various morphological features that are difficult to measure in the conventional analysis, and it is possible to use a combination of mitochondria filtering through features and subregion generation, which enables flexible analysis.

One limitation of this study is that the number of interactions may increase in longer analysis because the user parameters for image visualization should be frequently adjusted for error correction (e.g., transparency of the structure label should be high when the user wants to check raw image and should be low when the user performs correction). 
The expert also mentioned that it would be desirable to set up a few preferred visualizations in advance and switch between them. 

\begin{figure}
  \centering
  \includegraphics[width=0.7\linewidth]{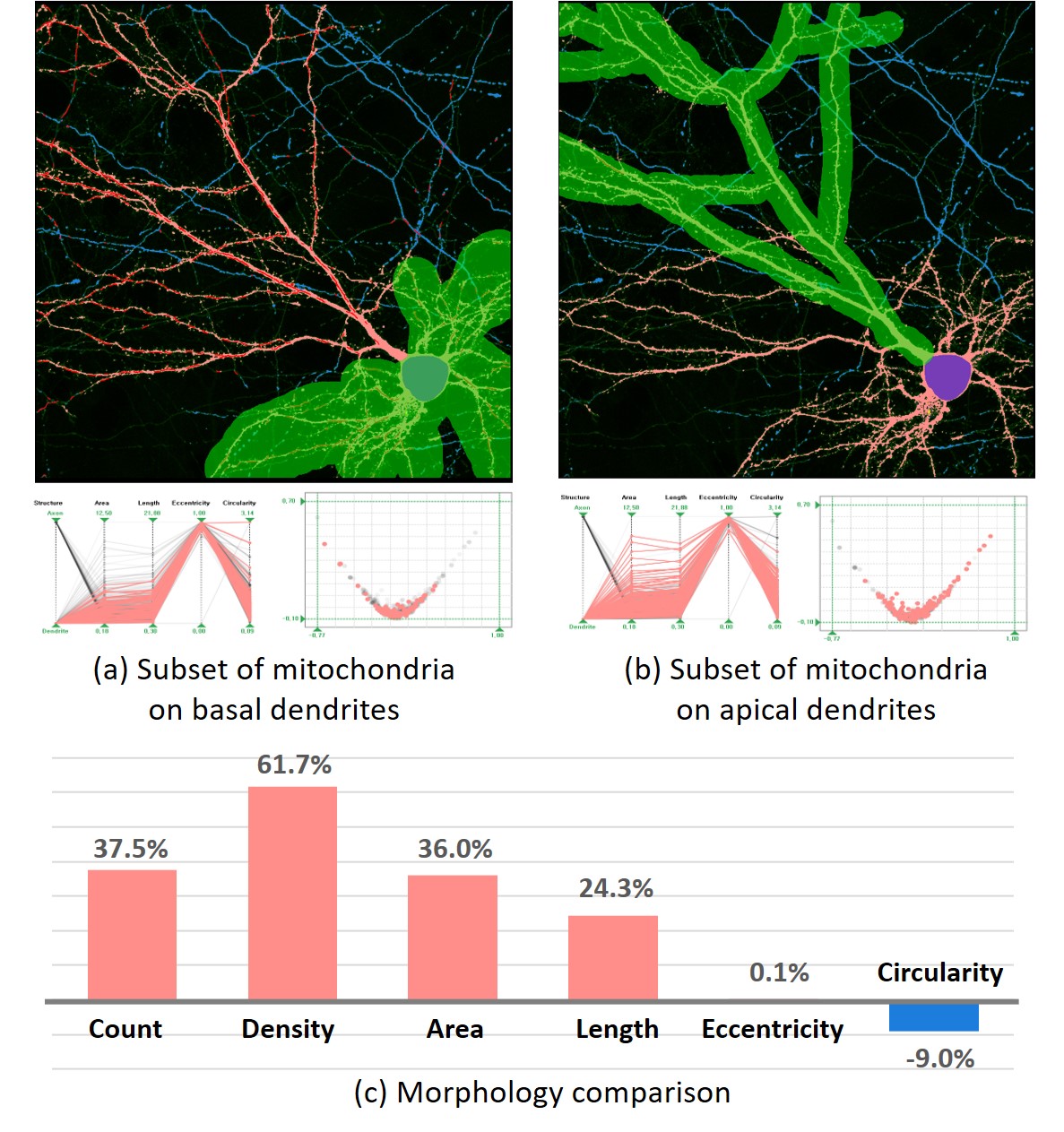}
  \caption{Example of morphology analysis for mitochondria on apical and basal dendrites. The apical and basal dendrites region can be created by the image-based subset generation function.}
  \label{fig:case2}
\end{figure}

\subsection{Further use cases}

We introduce three further use cases based on the expert feedback.

\textbf{Mitochondrial morphology and function in each area of neuron: }Since the image-based subset generation function of MitoVis can generate mitochondria subsets for specific regions of neurons, it is possible to perform mitochondria morphology analysis on detailed structures. One example is the analysis of apical and basal dendrites~\cite{wu2015differentiation}. Figure~\ref{fig:case2} presents an example of generating a mitochondria subset for apical and basal dendrites, and comparing morphology. Compared to basal dendrite, mitochondria located in apical dendrite tended to have longer and larger morphologies.

\textbf{Mitochondrial morphology differences between normal and disease models: }Several previous studies have attempted to discover the correlation between mitochondria and neurodegenerative disease models such as Alzheimer's disease~\cite{guo2020stabilization,du2010early}. MitoVis can help in this research field by analyzing changes in mitochondrial shape in disease models.

\textbf{Mitochondrial distribution and characteristics according to neuronal cell type: }The types of neurons are important for understanding the function of brain circuits~\cite{zeng2017neuronal}. If the mitochondria morphology of each type of neuron can be analyzed and defined through MitoVis, it will be possible to classify the neuron type.

\section{Discussion}
\label{sec:discussion}

\textbf{Contribution to the field of neuroscience: }MitoVis allows the automation of neuronal segmentation and mitochondrial analysis in each process; therefore, it saves an enormous amount of time for analyzing microscopy images of neuronal mitochondria. In particular, investigating the neuronal compartment-specific roles of mitochondria is becoming more important to understand neuronal function, and balancing the mitochondrial shape is critical.  For example, a recent study revealed that mitochondrial fission factor (MFF) deficiency elongates axonal mitochondria, and synaptic transmission and axon development are significantly impaired, although dendritic mitochondria are intact~\cite{lewis2018mff}. In addition, altered dendritic mitochondrial morphology has long been observed in various neurodegenerative disease patients and animal models~\cite{schon2011mitochondria,wang2009impaired,lee2018emerging,baek2017inhibition,ramonet2013optic}. Therefore, applying MitoVis for screening drugs will help find therapeutic candidates.

\begin{figure}
  \centering
  \includegraphics[width=0.7\linewidth]{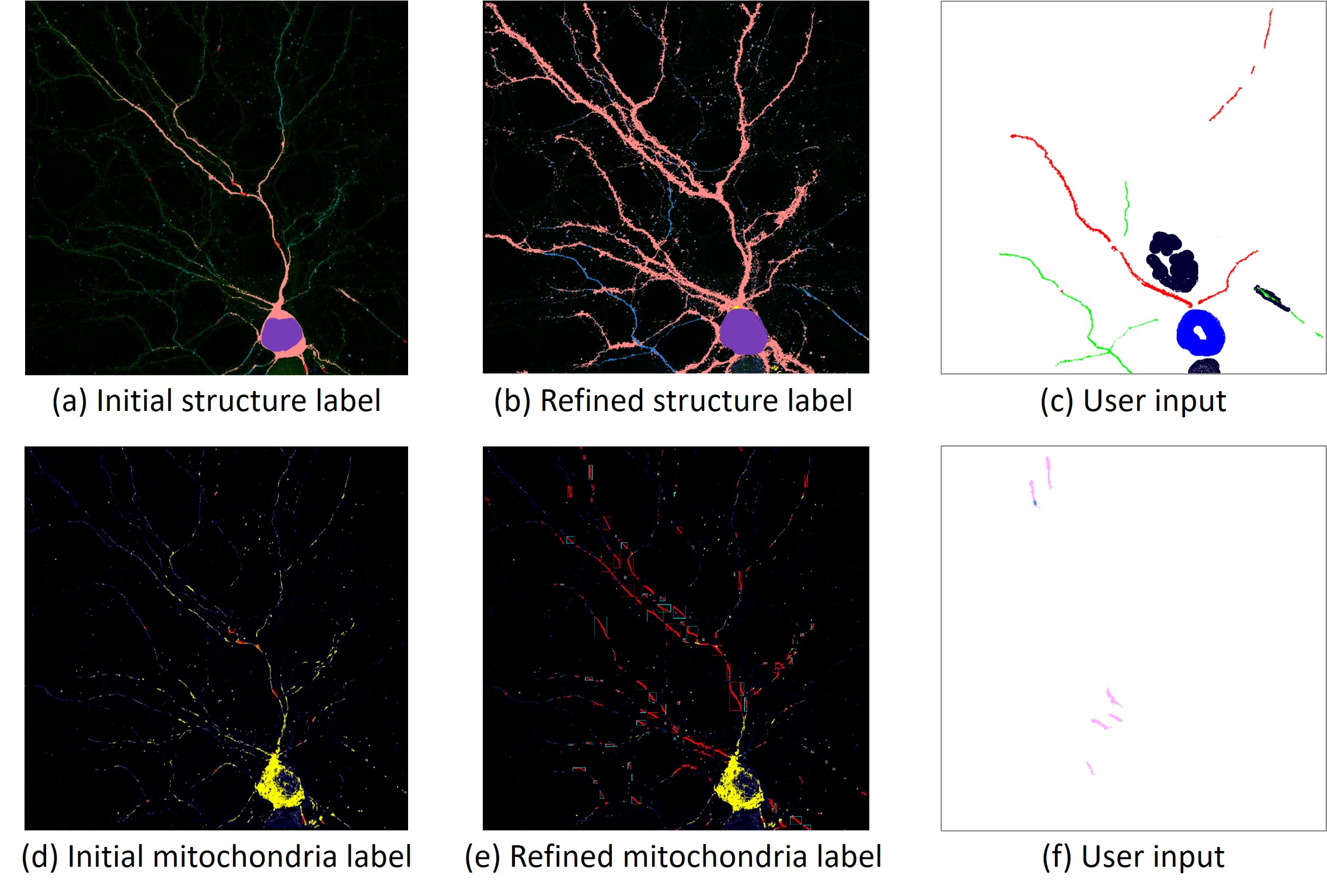}
  \caption{Use case of the interactive learning. Because this dataset is acquired without staining process, there are huge errors in the initial segmentation labels. With the interactive learning method, the error can be handled only with few user interactions.}
  \label{fig:fine_tuning}
\end{figure}

\textbf{Interactive learning approach: }In the case study, the expert manually corrected the errors and did not use the interactive learning method. 
We think this is because the number of errors was relatively small (i.e., initial prediction from the pre-trained model was acceptable) so the quick manual correction on several spots was sufficient and not too laborious. 
However, as shown in the dataset of Fig.~\ref{fig:fine_tuning} (image acquired without staining), if the pixel intensity distribution is significantly different from that of the training dataset, the initial inference may contain many errors. 
In such a case, manually correcting all the errors will be inefficient and laborious, and using the interactive learning approach can be effective and useful to reduce the manual effort.


\textbf{Limitation: }The image visualization of MitoVis is realized by manually adjusting various parameters. Because appropriate visualization is different depending on the task to be performed, there is a situation wherein parameters need to be adjusted frequently, which affects the usability. For example, to correct the structure label, the user had to repeatedly increase and decrease the transparency of the structure label to check the raw Venus image.
Visualization recommendations can be an approach to address these limitations. Several studies recommend visualization as a rule-based approach~\cite{wongsuphasawat2015voyager, mafrur2018dive} and a machine learning-based approach ~\cite{hu2019vizml}. 
However, research on a method that recommends appropriate image visualization according to the user's task has not been performed yet, which is left for future work.

\section{Conclusion and future work}

We introduced MitoVis, an interactive intelligent visual analytics system to perform neuronal mitochondria morphology analysis quickly and effectively. MitoVis enables rapid analysis by drastically reducing the time required for pre-processing through deep learning, and enables precise analysis through effective visualization, interaction and interactive learning approaches. 
In addition, it enables flexible analysis by allowing the generation of various mitochondria subsets using feature-based and image-based subset generation methods. 
These strengths of MitoVis are expected to enable the analysis of mitochondria morphology in various fields such as neurodegenerative disease models and neuronal cell types.

 In the future, we plan to conduct an in-depth analysis of disease models and neuronal cell types using MitoVis. 
 To further improve the usability of MitoVis, we plan to develop an advanced visualization recommendation system. 
 %

\newpage



\begin{thebibliography}{10}

\bibitem{baek2017inhibition}
Seung~Hyun Baek, So~Jung Park, Jae~In Jeong, Sung~Hyun Kim, Jihoon Han, Jae~Won
  Kyung, Sang-Ha Baik, Yuri Choi, Bo~Youn Choi, Jin~Su Park, et~al.
\newblock Inhibition of drp1 ameliorates synaptic depression, a$\beta$
  deposition, and cognitive impairment in an alzheimer's disease model.
\newblock {\em Journal of Neuroscience}, 37(20):5099--5110, 2017.

\bibitem{bro2014principal}
Rasmus Bro and Age~K Smilde.
\newblock Principal component analysis.
\newblock {\em Analytical methods}, 6(9):2812--2831, 2014.

\bibitem{chicurel1992three}
Marina~E Chicurel and Kristen~M Harris.
\newblock Three-dimensional analysis of the structure and composition of ca3
  branched dendritic spines and their synaptic relationships with mossy fiber
  boutons in the rat hippocampus.
\newblock {\em Journal of comparative neurology}, 325(2):169--182, 1992.

\bibitem{dickey2011pka}
Audrey~S Dickey and Stefan Strack.
\newblock Pka/akap1 and pp2a/b$\beta$2 regulate neuronal morphogenesis via drp1
  phosphorylation and mitochondrial bioenergetics.
\newblock {\em Journal of Neuroscience}, 31(44):15716--15726, 2011.

\bibitem{du2010early}
Heng Du, Lan Guo, Shiqiang Yan, Alexander~A Sosunov, Guy~M McKhann, and
  Shirley~ShiDu Yan.
\newblock Early deficits in synaptic mitochondria in an alzheimer's disease
  mouse model.
\newblock {\em Proceedings of the National Academy of Sciences},
  107(43):18670--18675, 2010.

\bibitem{fecher2019cell}
Caroline Fecher, Laura Trov{\`o}, Stephan~A M{\"u}ller, Nicolas Snaidero,
  Jennifer Wettmarshausen, Sylvia Heink, Oskar Ortiz, Ingrid Wagner, Ralf
  K{\"u}hn, Jana Hartmann, et~al.
\newblock Cell-type-specific profiling of brain mitochondria reveals functional
  and molecular diversity.
\newblock {\em Nature neuroscience}, 22(10):1731--1742, 2019.

\bibitem{feng2015neutube}
Linqing Feng, Ting Zhao, and Jinhyun Kim.
\newblock neutube 1.0: a new design for efficient neuron reconstruction
  software based on the swc format.
\newblock {\em eneuro}, 2(1), 2015.

\bibitem{fischer2020mitosegnet}
Christian~A Fischer, Laura Besora-Casals, St{\'e}phane~G Rolland, Simon
  Haeussler, Kritarth Singh, Michael Duchen, Barbara Conradt, and Carsten Marr.
\newblock Mitosegnet: Easy-to-use deep learning segmentation for analyzing
  mitochondrial morphology.
\newblock {\em Iscience}, 23(10):101601, 2020.

\bibitem{fogo2021machine}
Garrett~M Fogo, Anthony~R Anzell, Kathleen~J Maheras, Sarita Raghunayakula,
  Joseph~M Wider, Katlynn~J Emaus, Timothy~D Bryson, Melissa~J Bukowski,
  Robert~W Neumar, Karin Przyklenk, et~al.
\newblock Machine learning-based classification of mitochondrial morphology in
  primary neurons and brain.
\newblock {\em Scientific reports}, 11(1):1--12, 2021.

\bibitem{guo2020stabilization}
Baihong Guo, Yangmei Huang, Qingtao Gao, and Qiang Zhou.
\newblock Stabilization of microtubules improves cognitive functions and axonal
  transport of mitochondria in alzheimer's disease model mice.
\newblock {\em Neurobiology of Aging}, 96:223--232, 2020.

\bibitem{hu2019vizml}
Kevin Hu, Michiel~A Bakker, Stephen Li, Tim Kraska, and C{\'e}sar Hidalgo.
\newblock Vizml: A machine learning approach to visualization recommendation.
\newblock In {\em Proceedings of the 2019 CHI Conference on Human Factors in
  Computing Systems}, pages 1--12, 2019.

\bibitem{jang2019interactive}
Won-Dong Jang and Chang-Su Kim.
\newblock Interactive image segmentation via backpropagating refinement scheme.
\newblock In {\em Proceedings of the IEEE/CVF Conference on Computer Vision and
  Pattern Recognition}, pages 5297--5306, 2019.

\bibitem{kasthuri2015saturated}
Narayanan Kasthuri, Kenneth~Jeffrey Hayworth, Daniel~Raimund Berger,
  Richard~Lee Schalek, Jos{\'e}~Angel Conchello, Seymour Knowles-Barley, Dongil
  Lee, Amelio V{\'a}zquez-Reina, Verena Kaynig, Thouis~Raymond Jones, et~al.
\newblock Saturated reconstruction of a volume of neocortex.
\newblock {\em Cell}, 162(3):648--661, 2015.

\bibitem{kim2020altered}
Gyu~Hyun Kim, Yinhua Zhang, Hyae~Rim Kang, Seung-Hyun Lee, Jiwon Shin, Chan~Hee
  Lee, Hyojin Kang, Ruiying Ma, Chunmei Jin, Yoonhee Kim, et~al.
\newblock Altered presynaptic function and number of mitochondria in the medial
  prefrontal cortex of adult cyfip2 heterozygous mice.
\newblock {\em Molecular brain}, 13(1):1--5, 2020.

\bibitem{lee2018emerging}
Annie Lee, Yusuke Hirabayashi, Seok-Kyu Kwon, Tommy~L Lewis~Jr, and Franck
  Polleux.
\newblock Emerging roles of mitochondria in synaptic transmission and
  neurodegeneration.
\newblock {\em Current opinion in physiology}, 3:82--93, 2018.

\bibitem{lewis2018mff}
Tommy~L Lewis, Seok-Kyu Kwon, Annie Lee, Reuben Shaw, and Franck Polleux.
\newblock Mff-dependent mitochondrial fission regulates presynaptic release and
  axon branching by limiting axonal mitochondria size.
\newblock {\em Nature communications}, 9(1):1--15, 2018.

\bibitem{li2004importance}
Zheng Li, Ken-Ichi Okamoto, Yasunori Hayashi, and Morgan Sheng.
\newblock The importance of dendritic mitochondria in the morphogenesis and
  plasticity of spines and synapses.
\newblock {\em Cell}, 119(6):873--887, 2004.

\bibitem{lihavainen2012mytoe}
Eero Lihavainen, Jarno M{\"a}kel{\"a}, Johannes~N Spelbrink, and Andre~S
  Ribeiro.
\newblock Mytoe: automatic analysis of mitochondrial dynamics.
\newblock {\em Bioinformatics}, 28(7):1050--1051, 2012.

\bibitem{mafrur2018dive}
Rischan Mafrur, Mohamed~A Sharaf, and Hina~A Khan.
\newblock Dive: Diversifying view recommendation for visual data exploration.
\newblock In {\em Proceedings of the 27th ACM International Conference on
  Information and Knowledge Management}, pages 1123--1132, 2018.

\bibitem{magliaro2017manual}
Chiara Magliaro, Alejandro~L Callara, Nicola Vanello, and Arti Ahluwalia.
\newblock A manual segmentation tool for three-dimensional neuron datasets.
\newblock {\em Frontiers in neuroinformatics}, 11:36, 2017.

\bibitem{mcinnes2018umap}
Leland McInnes, John Healy, and James Melville.
\newblock Umap: Uniform manifold approximation and projection for dimension
  reduction.
\newblock {\em arXiv preprint arXiv:1802.03426}, 2018.

\bibitem{merrill2017measuring}
Ronald~A Merrill, Kyle~H Flippo, and Stefan Strack.
\newblock Measuring mitochondrial shape with imagej.
\newblock In {\em Techniques to investigate mitochondrial function in neurons},
  pages 31--48. Springer, 2017.

\bibitem{chan2021noiseloss}
Chanmin Park, Kanggeun Lee, Suyeon Kim, Fatma Sema~Canbakis Cecen, Seok-Kyu
  Kwon, and Won-Ki Jeong.
\newblock Neuron segmentation using incomplete and noisy labels via adaptive
  learning with structure priors.
\newblock In {\em 2021 IEEE 18th International Symposium on Biomedical Imaging
  (ISBI 2021)}. IEEE, 2021.

\bibitem{paszke2019pytorch}
Adam Paszke, Sam Gross, Francisco Massa, Adam Lerer, James Bradbury, Gregory
  Chanan, Trevor Killeen, Zeming Lin, Natalia Gimelshein, Luca Antiga, et~al.
\newblock Pytorch: An imperative style, high-performance deep learning library.
\newblock {\em arXiv preprint arXiv:1912.01703}, 2019.

\bibitem{peng2014extensible}
Hanchuan Peng, Alessandro Bria, Zhi Zhou, Giulio Iannello, and Fuhui Long.
\newblock Extensible visualization and analysis for multidimensional images
  using vaa3d.
\newblock {\em Nature protocols}, 9(1):193--208, 2014.

\bibitem{popov2005mitochondria}
Victor Popov, Nikolai~I Medvedev, Heather~A Davies, and Michael~G Stewart.
\newblock Mitochondria form a filamentous reticular network in hippocampal
  dendrites but are present as discrete bodies in axons: A three-dimensional
  ultrastructural study.
\newblock {\em Journal of Comparative Neurology}, 492(1):50--65, 2005.

\bibitem{ramonet2013optic}
D~Ramonet, C~Perier, A~Recasens, B~Dehay, J~Bove, V~Costa, L~Scorrano, and
  M~Vila.
\newblock Optic atrophy 1 mediates mitochondria remodeling and dopaminergic
  neurodegeneration linked to complex i deficiency.
\newblock {\em Cell Death \& Differentiation}, 20(1):77--85, 2013.

\bibitem{rangaraju2019spatially}
Vidhya Rangaraju, Marcel Lauterbach, and Erin~M Schuman.
\newblock Spatially stable mitochondrial compartments fuel local translation
  during plasticity.
\newblock {\em Cell}, 176(1-2):73--84, 2019.

\bibitem{rangaraju2019pleiotropic}
Vidhya Rangaraju, Tommy~L Lewis, Yusuke Hirabayashi, Matteo Bergami, Elisa
  Motori, Romain Cartoni, Seok-Kyu Kwon, and Julien Courchet.
\newblock Pleiotropic mitochondria: the influence of mitochondria on neuronal
  development and disease.
\newblock {\em Journal of Neuroscience}, 39(42):8200--8208, 2019.

\bibitem{sardar2020iris}
Mousumi Sardar, Subhashis Banerjee, and Sushmita Mitra.
\newblock Iris segmentation using interactive deep learning.
\newblock {\em IEEE Access}, 8:219322--219330, 2020.

\bibitem{schon2011mitochondria}
Eric~A Schon and Serge Przedborski.
\newblock Mitochondria: the next (neurode) generation.
\newblock {\em Neuron}, 70(6):1033--1053, 2011.

\bibitem{sun2019meta}
Qianru Sun, Yaoyao Liu, Tat-Seng Chua, and Bernt Schiele.
\newblock Meta-transfer learning for few-shot learning.
\newblock In {\em Proceedings of the IEEE/CVF Conference on Computer Vision and
  Pattern Recognition}, pages 403--412, 2019.

\bibitem{varkuti2020neuron}
Boglarka~H Varkuti, Miklos Kepiro, Ze~Liu, Kyle Vick, Yosef Avchalumov, Rodrigo
  Pacifico, Courtney~M MacMullen, Theodore~M Kamenecka, Sathyanarayanan~V
  Puthanveettil, and Ronald~L Davis.
\newblock Neuron-based high-content assay and screen for cns active
  mitotherapeutics.
\newblock {\em Science advances}, 6(2):eaaw8702, 2020.

\bibitem{wang2018interactive}
Guotai Wang, Wenqi Li, Maria~A Zuluaga, Rosalind Pratt, Premal~A Patel, Michael
  Aertsen, Tom Doel, Anna~L David, Jan Deprest, S{\'e}bastien Ourselin, et~al.
\newblock Interactive medical image segmentation using deep learning with
  image-specific fine tuning.
\newblock {\em IEEE transactions on medical imaging}, 37(7):1562--1573, 2018.

\bibitem{wang2009impaired}
Xinglong Wang, BO~Su, Hyoung-gon Lee, Xinyi Li, George Perry, Mark~A Smith, and
  Xiongwei Zhu.
\newblock Impaired balance of mitochondrial fission and fusion in alzheimer's
  disease.
\newblock {\em Journal of neuroscience}, 29(28):9090--9103, 2009.

\bibitem{wiemerslage2016quantification}
Lyle Wiemerslage and Daewoo Lee.
\newblock Quantification of mitochondrial morphology in neurites of
  dopaminergic neurons using multiple parameters.
\newblock {\em Journal of neuroscience methods}, 262:56--65, 2016.

\bibitem{wongsuphasawat2015voyager}
Kanit Wongsuphasawat, Dominik Moritz, Anushka Anand, Jock Mackinlay, Bill Howe,
  and Jeffrey Heer.
\newblock Voyager: Exploratory analysis via faceted browsing of visualization
  recommendations.
\newblock {\em IEEE transactions on visualization and computer graphics},
  22(1):649--658, 2015.

\bibitem{wu2015differentiation}
You~Kure Wu, Kazuto Fujishima, and Mineko Kengaku.
\newblock Differentiation of apical and basal dendrites in pyramidal cells and
  granule cells in dissociated hippocampal cultures.
\newblock {\em PloS one}, 10(2):e0118482, 2015.

\bibitem{zahedi2018deep}
Atena Zahedi, Vincent On, Rattapol Phandthong, Angela Chaili, Guadalupe Remark,
  Bir Bhanu, and Prue Talbot.
\newblock Deep analysis of mitochondria and cell health using machine learning.
\newblock {\em Scientific reports}, 8(1):1--15, 2018.

\bibitem{zeng2017neuronal}
Hongkui Zeng and Joshua~R Sanes.
\newblock Neuronal cell-type classification: challenges, opportunities and the
  path forward.
\newblock {\em Nature Reviews Neuroscience}, 18(9):530--546, 2017.

\bibitem{zhou2020gtree}
Hang Zhou, Shiwei Li, Anan Li, Qing Huang, Feng Xiong, Ning Li, Jiacheng Han,
  Hongtao Kang, Yijun Chen, Yun Li, et~al.
\newblock Gtree: an open-source tool for dense reconstruction of brain-wide
  neuronal population.
\newblock {\em Neuroinformatics}, pages 1--13, 2020.

\end{thebibliography}





\end{document}